\documentclass[10pt,journal,compsoc]{IEEEtran}
\ifCLASSOPTIONcompsoc
  \usepackage[compress]{cite}
\else
  \usepackage{cite}
\fi

\usepackage{ragged2e}
\usepackage{underscore}
\usepackage{adjustbox}
\usepackage{hyperref}
\usepackage{amssymb}
\usepackage{graphicx}
\usepackage{multirow}
\usepackage{bbding}
\usepackage{float}
\usepackage{cancel}
\usepackage{url}
\usepackage{graphicx}
\usepackage{color}
\usepackage{booktabs}
\usepackage[linesnumbered, ruled]{algorithm2e}
\usepackage{algorithm2e,setspace}
\usepackage{stfloats}
\usepackage{multirow}
\usepackage{makecell}
\usepackage{multicol}
\usepackage[noend]{algpseudocode}
\usepackage{bm}
\usepackage{float}
\usepackage[table,xcdraw]{xcolor}
\usepackage[normalem]{ulem}

\useunder{\uline}{\ul}{}
\usepackage{float}  
\usepackage{lipsum}


\ifCLASSINFOpdf
\else
\fi
\begin{document}
%
% paper title
% Titles are generally capitalized except for words such as a, an, and, as,
% at, but, by, for, in, nor, of, on, or, the, to and up, which are usually
% not capitalized unless they are the first or last word of the title.
% Linebreaks \\ can be used within to get better formatting as desired.
% Do not put math or special symbols in the title.
%\title{Knowledge Distillation Meets Federated Multi-task Learning in Mobile Edge Computing}

%\title{FedCache: A Novel Federated Learning Architecture for Personalized Edge Intelligence}
%\title{Personalized Edge Intelligence with Cache-based Knowledge Retrieval}
%\title{FedCache: A Novel Federated Learning Architecture for Personalized Edge Intelligence}
%\title{Federated Learning with Server-Side Knowledge Caching and Retrieval: An Architecture for Personalized Edge Intelligence}
%\title{FedCache: A Novel Personalized Federated Learning Architecture with Knowledge Caching and Retrieval in the Edge}
\title{FedCache 2.0: Federated Edge Learning with Knowledge Caching and Dataset Distillation}
%
%
% author names and IEEE memberships
% note positions of commas and nonbreaking spaces ( ~ ) LaTeX will not break
% a structure at a ~ so this keeps an author's name from being broken across
% two lines.
% use \thanks{} to gain access to the first footnote area
% a separate \thanks must be used for each paragraph as LaTeX2e's \thanks
% was not built to handle multiple paragraphs
%
%
%\IEEEcompsocitemizethanks is a special \thanks that produces the bulleted
% lists the Computer Society journals use for "first footnote" author
% affiliations. Use \IEEEcompsocthanksitem which works much like \item
% for each affiliation group. When not in compsoc mode,
% \IEEEcompsocitemizethanks becomes like \thanks and
% \IEEEcompsocthanksitem becomes a line break with idention. This
% facilitates dual compilation, although admittedly the differences in the
% desired content of \author between the different types of papers makes a
% one-size-fits-all approach a daunting prospect. For instance, compsoc 
% journal papers have the author affiliations above the "Manuscript
% received ..."  text while in non-compsoc journals this is reversed. Sigh.
\author{Quyang~Pan,
         Sheng~Sun,
         Zhiyuan~Wu,~\IEEEmembership{Member,~IEEE,}
         Yuwei~Wang,~\IEEEmembership{Member,~IEEE,}
         Min~Liu,~\IEEEmembership{Senior~Member,~IEEE,}
         ~Bo~Gao,~\IEEEmembership{Member,~IEEE},
         and~Jingyuan~Wang,
\IEEEcompsocitemizethanks{
\IEEEcompsocthanksitem Quyang Pan and Zhiyuan Wu are with the State Key Laboratory of Processors, Institute of Computing Technology, Chinese Academy of Sciences, Beijing, China, and also with the University of Chinese Academy of Sciences, Beijing, China.
E-mails: \{panquyang23s, wuzhiyuan22s\}@ict.ac.cn. %tianliu.he@foxmail.com.
\IEEEcompsocthanksitem Sheng Sun and Yuwei Wang are with the State Key Laboratory of Processors, Institute of Computing Technology, Chinese Academy of Sciences, Beijing, China.
E-mails: \{sunsheng, ywwang\}@ict.ac.cn.
\IEEEcompsocthanksitem Min Liu is with the State Key Laboratory of Processors, Institute of Computing Technology, Chinese Academy of Sciences, Beijing, China, and also with the Zhongguancun Laboratory, Beijing, China.
E-mail: liumin@ict.ac.cn
\IEEEcompsocthanksitem Bo Gao is with the School of Computer and Information Technology, and the Engineering Research Center of Network Management Technology for High-Speed Railway of Ministry of Education, Beijing Jiaotong University, Beijing, China.
E-mail: bogao@bjtu.edu.cn.
\IEEEcompsocthanksitem Jingyuan Wang is with the School of Computer Science and Engineering, Beihang Unversity, Beijing, China.
E-mail: jywang@buaa.edu.cn.
\IEEEcompsocthanksitem Corresponding author: Yuwei Wang.
}% <-this % stops an unwanted space
\thanks{
	This work was supported by the National Key Research and Development Program of China (No. 2023YFB2703701), the National Natural Science Foundation of China (No. 62472410, No. 62072436).}
}
% note the % following the last \IEEEmembership and also \thanks - 
% these prevent an unwanted space from occurring between the last author name
% and the end of the author line. i.e., if you had this:
% 
% \author{....lastname \thanks{...} \thanks{...} }
%                     ^------------^------------^----Do not want these spaces!
%
% a space would be appended to the last name and could cause every name on that
% line to be shifted left slightly. This is one of those "LaTeX things". For
% instance, "\textbf{A} \textbf{B}" will typeset as "A B" not "AB". To get
% "AB" then you have to do: "\textbf{A}\textbf{B}"
% \thanks is no different in this regard, so shield the last } of each \thanks
% that ends a line with a % and do not let a space in before the next \thanks.
% Spaces after \IEEEmembership other than the last one are OK (and needed) as
% you are supposed to have spaces between the names. For what it is worth,
% this is a minor point as most people would not even notice if the said evil
% space somehow managed to creep in.

% The paper headers
\markboth{Under Review}%
% \markboth{{\tiny This work has been submitted to the IEEE for possible publication. Copyright may be transferred without notice, after which this version may no longer be accessible. }}
{Shell \MakeLowercase{\textit{et al.}}: Bare Demo of IEEEtran.cls for Computer Society Journals}
% The only time the second header will appear is for the odd numbered pages
% after the title page when using the twoside option.
% 
% *** Note that you probably will NOT want to include the author's ***
% *** name in the headers of peer review papers.                   ***
% You can use \ifCLASSOPTIONpeerreview for conditional compilation here if
% you desire.

% The publisher's ID mark at the bottom of the page is less important with
% Computer Society journal papers as those publications place the marks
% outside of the main text columns and, therefore, unlike regular IEEE
% journals, the available text space is not reduced by their presence.
% If you want to put a publisher's ID mark on the page you can do it like
% this:
%\IEEEpubid{0000--0000/00\$00.00~\copyright~2015 IEEE}
% or like this to get the Computer Society new two part style.
%\IEEEpubid{\makebox[\columnwidth]{\hfill 0000--0000/00/\$00.00~\copyright~2015 IEEE}%
%\hspace{\columnsep}\makebox[\columnwidth]{Published by the IEEE Computer Society\hfill}}
% Remember, if you use this you must call \IEEEpubidadjcol in the second
% column for its text to clear the IEEEpubid mark (Computer Society jorunal
% papers don't need this extra clearance.)

% use for special paper notices
%\IEEEspecialpapernotice{(Invited Paper)}

% for Computer Society papers, we must declare the abstract and index terms
% PRIOR to the title within the \IEEEtitleabstractindextext IEEEtran
% command as these need to go into the title area created by \maketitle.
% As a general rule, do not put math, special symbols or citations
% in the abstract or keywords.

\IEEEtitleabstractindextext{%
\begin{abstract}
\justifying
Federated Edge Learning (FEL) has emerged as a promising approach for enabling edge devices to collaboratively train machine learning models while preserving data privacy. Despite its advantages, practical FEL deployment faces significant challenges stemming from device constraints and device-server interactions, necessitating heterogeneous, user-adaptive model training with limited and uncertain communication conditions. \textcolor{black}{Knowledge Cache-driven Federated Learning (FedCache) is a promising architecture that enables communication-efficient and heterogeneous-aware collaborative training in edge computing scenarios. However, previous work is limited by the intrinsic nature of logits-based interactions, leading to performance bottlenecks due to the poor richness of exchanged information for on-device model optimization. To tackle this issue, we introduce FedCache 2.0, a novel personalized FEL architecture that enhances the exchange of optimization insights while delivering state-of-the-art performance with efficient communication.} FedCache 2.0 incorporates the benefits of both dataset distillation and knowledge cache-driven federated learning by storing and organizing distilled data as knowledge in the server-side knowledge cache, \textcolor{black}{allowing devices to periodically download and utilize personalized knowledge for local model optimization.} Moreover, a device-centric cache sampling strategy is introduced to tailor transferred knowledge for individual devices within controlled communication bandwidth. Extensive experiments on five datasets covering image recognition, audio understanding, and mobile sensor data mining tasks demonstrate that (1)  FedCache 2.0 significantly outperforms state-of-the-art methods regardless of model structures, data distributions, and modalities. (2) FedCache 2.0 can train splendid personalized on-device models with at least $\times$28.6 improvement in communication efficiency. Our code is available at \textit{\url{https://github.com/poppanda/FedCache2.0}}.
\end{abstract}
}
% make the title area
\maketitle

% To allow for easy dual compilation without having to reenter the
% abstract/keywords data, the \IEEEtitleabstractindextext text will
% not be used in maketitle, but will appear (i.e., to be "transported")
% here as \IEEEdisplaynontitleabstractindextext when the compsoc 
% or transmag modes are not selected <OR> if conference mode is selected 
% - because all conference papers position the abstrac like regular
% papers do.
\IEEEdisplaynontitleabstractindextext
% \IEEEdisplaynontitleabstractindextext has no effect when using
% compsoc or transmag under a non-conference mode.

% For peer review papers, you can put extra information on the cover
% page as needed:
% \ifCLASSOPTIONpeerreview
% \begin{center} \bfseries EDICS Category: 3-BBND \end{center}
% \fi
%
% For peerreview papers, this IEEEtran command inserts a page break and
% creates the second title. It will be ignored for other modes.

\IEEEpeerreviewmaketitle
\vspace{20pt}
\IEEEraisesectionheading{\section{Introduction}}
\IEEEPARstart{F}ederated Edge Learning (FEL) \cite{tak2021federated,duan2023combining} \textcolor{black}{is a specialized form of Federated Learning \cite{yang2019federated} designed to operate at the edge of the network. It enables edge devices (clients) to jointly train machine learning models under the coordination of an edge server (server) without sharing raw data.} With the growing prevalence of mobile and Internet of Things (IoT) devices coupled with increasing concerns over data privacy, FEL has empowered wide adoption of various on-device Artificial Intelligence (AI) applications, including smart transportation \cite{transportation}, healthcare \cite{healthcare}, and recommendation \cite{yuan2023federated,guo2021prefer}.

\color{black}
Despite its promising potential, FEL faces significant challenges in practical deployment. One of the primary challenges is the diversity of edge device infrastructures. Devices such as smartwatches, mobile phones, and tablets differ significantly in terms of computational power, storage capacity, and battery life \cite{tak2021federated,yu2021toward}. This diversity calls for the deployment of highly scalable, user-adaptive models that can accommodate heterogeneous hardware specifications across devices \cite{tak2021federated,duan2023combining,pflsurvey,yu2021toward}. Even among devices with similar hardware configurations, data distributions and user preferences are often highly individualized, further complicating model generalization \cite{tan2022towards,mills2021multi,wu2023fedict}. Another challenge arises from the communication limitations in mobile edge networks. Edge devices frequently operate under low bandwidth and unreliable network conditions \cite{tak2021federated,zhang2022federated,zhu2022resilient}, making the transmission of large-scale model parameters impractical due to the precious nature of wireless channels \cite{al2023edge,wu2022communication}. Moreover, devices are often intermittently online \cite{zhu2022resilient}, further complicating the coordination required for collaborative model updates. The aforementioned constraints hinder the efficiency of the FEL process, ultimately impacting the systems' overall performance and the benefits it delivers to users. 
\color{black}

\begin{figure}[t]
    \centering
    \begin{minipage}{\linewidth}
        \centering
        \includegraphics[width=0.75\textwidth]{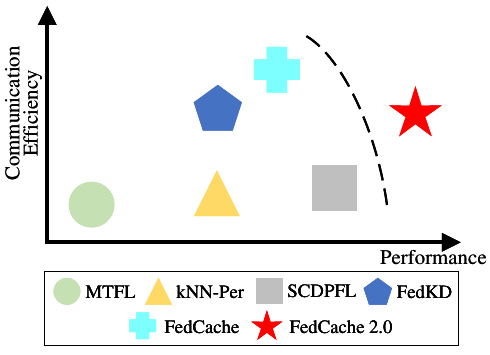}
        \caption{\textcolor{black}{Comparison of FedCache 2.0 with state-of-the-art methods in terms of performance and communication efficiency. FedCache 2.0 demonstrates significant performance improvements over state-of-the-art methods with substantially reduced communication overhead compared with model aggregation-based FL methods (MTFL \cite{mills2021multi}, kNN-Per \cite{marfoq2022personalized}, and SCDPFL\cite{chen2024spectral}), as indicated by its position in the top-right of the figure.}}
        \label{compare-fig}
    \end{minipage}\par\medskip
    
\end{figure}

Numerous studies have sought to overcome the aforementioned challenges in FEL. Heterogeneous FL \cite{zhu2021data,li2019fedmd,wu2024exploring} focuses on enabling collaborative training across devices with different model architectures, allowing adaptation to varying computational resources and hardware capabilities. Personalized or multi-task FL \cite{mills2021multi,marfoq2022personalized,jin2022personalized} centers on developing models tailored to individual devices, accommodating diverse user preferences. Communication-efficient FL \cite{wu2022communication,sattler2021cfd,hou2023efficient} typically incorporates techniques such as model compression or quantization, reducing the training communication burden with little performance loss. \textcolor{black}{However, each of these approaches addresses only a single challenge}, limiting their applicability in real-world edge deployments where multiple challenges coexist.

%改到此处
Given the complexity of the aforementioned challenges, it is evident that a multi-faceted FEL solution is urgently needed. Fortunately, knowledge cache-driven FL (FedCache) \cite{wu2024fedcache} offers a prevailing paradigm that revolutionizes the mainstream parameters interaction protocol \cite{mcmahan2017communication,li2020federated,reddi2021adaptive,spectral,mills2021multi,jin2022personalized,lee2022preservation,wu2024federated} in FEL. \textcolor{black}{By performing on-device personalized distillation driven by the self-organizing server-side knowledge cache, FedCache facilitates communication-efficient and heterogeneous-compatible collaborative training without requiring multiple devices to remain online simultaneously.} However, the performance of FedCache is limited by the logits interaction design, which restricts the amount and quality of information that can be interacted between devices and the server. In addition, the applicability of FedCache across various data modalities and application tasks is also restricted due to its reliance on task-specific encoders.
%\textcolor{black}{Therefore, there remains a pressing need for an FEL method that can adapt to the inherent challenges of edge environments while delivering higher-performance models to users.}
\color{black}

In this paper, we introduce FedCache 2.0, a novel personalized FEL architecture that improves the performance of heterogeneous on-device models with efficient communication and uncertain connection tolerance, \textcolor{black}{as displayed in Fig. \ref{compare-fig}.} \textcolor{black}{Specifically, FedCache 2.0 revolute the transferred knowledge by shifting from logits to distilled data \cite{lei2023comprehensive}, offering a novel interaction paradigm between devices and the server.} In our novel design, devices perform dataset distillation with the assistance of cached knowledge from the remote server. The distilled data is then shared with the server, ensuring the knowledge cache remains updated with the latest information. To balance system performance and communication efficiency, a device-centric cache sampling strategy is proposed for tailoring transferred knowledge for individual devices within the constraints of available communication bandwidth. 
The key superiorities of FedCache 2.0 compared with the original FedCache are twofold. First, FedCache 2.0 provides richer information characterization capabilities by storing and transferring distilled synthetic data rather than logits, enabling on-device models to optimize with sufficient server-side information and achieve better precision. Second, FedCache 2.0 adopts a more generalized data anonymization method, enhancing its extensibility to a broader range of data modalities and application tasks. Our proposed architecture maintains the advantages of model heterogeneity allowance, learning personalization, uncertain connection tolerance, and efficient communication from FedCache, while also achieving remarkable performance gains by fully exploiting the knowledge from distilled data.

%\textbf{Contributions.} 
The main contributions of this paper are as follows: 
\begin{itemize}
    \item 
    \textcolor{black}{We propose FedCache 2.0, a novel personalized FEL architecture that integrates dataset distillation with knowledge cache-driven federated learning. \textbf{To our best knowledge, FedCache 2.0 is the first federated learning architecture to achieve state-of-the-art performance while tackling the multifaceted challenges of edge deployment.}}
    
    \item 
    \textcolor{black}{We introduce federated dataset distillation and device-centric cache sampling that matches FedCache 2.0 design, facilitating knowledge generation, storage, and organization as well as personalized model training, with data privacy protected.}
    
    \item 
    We conduct comprehensive experiments on five datasets, encompassing image recognition, audio understanding, and mobile sensor data mining tasks. Built upon diversified data heterogeneity, model settings, and application scenarios, FedCache 2.0 not only consistently outperforms state-of-the-art methods (at least 1.7\% average User model Accuracy \cite{mills2021multi} enhancement) in all considered settings, but also achieves better communication efficiency (at least $\times$29.6) compared with baseline algorithms.
\end{itemize}

%My paper is to be submitted to NeurIPS. please help me polish the Preliminary part of my uploaded paper. Refer to the content of the whole paper.

\begin{figure*}[t]
        %\captionsetup{skip=4pt}
	\centering
	\includegraphics[width=1.0\textwidth]{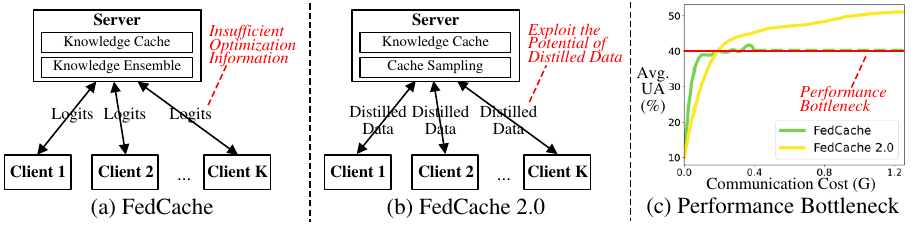}
	\caption{Comparison of FedCache and FedCache2.0. Results in (c) are derived on the CIFAR-10 dataset, taking $\alpha=0.5$ and $K=100$.}
        \label{limitation}
\end{figure*}

\section{Problem Statement and Reformulation}
\subsection{Background and Preliminary}
The main notations of this paper are summarized in Table \ref{notations-table}. Without loss of generality, we consider an FL system deployed at the edge of the network, comprising $K$ participating edge devices (clients) coordinated by an edge server (server). Each client $k \in\{1,2,... ,K\}$ owns its local dataset $\mathcal{D}^k=\bigcup\limits_{i = 1}^{|\mathcal{D}^k|} {\{ ({X_i^k},{y_i^k})\} }$ with $ |{\mathcal{D}^k}|$ samples, where each sample are with $D$ data dimensions and belong to one of $C$ distinct classes. Due to differentiated user behaviors, both the local training an testing datasets among clients are non-independently and identically distributed. Throughout this paper, the terms 'device' and 'client' are used interchangeably. Assume that the personalized model parameters of client $k$ are denoted as $W^k\in \mathbb{R}^{d^k}$, where $d^k$ indicates the number of parameters in the model of client $k$. Due to system heterogeneity among devices, the required model sizes may vary across clients, such that $d_l \neq d_m, \exists l,m\in\{1,2,... ,K\}$. Each client $k$ has a local objective $\mathcal{L}^i:\mathbb{R}^D\rightarrow \mathbb{R}$, which relies on its corresponding local data distribution. The overall goal is to minimize the expected objective across all clients, which is formally expressed as:

\label{main-notations}
    \begin{table}[t]
    \centering
    \caption{Main notations with descriptions.}
    \renewcommand\arraystretch{0.9}
    \begin{tabular}{>{\centering\arraybackslash}m{1.0cm} >{\centering\arraybackslash}m{6.5cm}}
        \toprule
        \textbf{Notation} & \textbf{Description} \\ 
        \midrule
        $K$ & Number of participating edge devices (clients) \\
    $\mathcal{D}^k$ & Local dataset of client $k$ \\
    $(X_i^k, y_i^k)$ & The $i$-th sample in the local dataset of client $k$ \\
    $C$ & Number of classes \\
    $W^k$ & Personalized model parameters of client $k$ \\
    $d^k$ & Number of parameters in the model of client $k$ \\
    $\mathcal{L}_i$ & Local objective function of client $k$ \\
    $KC$ & Knowledge cache on the server \\
    $\mathcal{L}_{CE}$ & Cross-entropy loss \\
    $\varphi$ & Softmax function \\
    $F^k$ & Prediction function of the model on client $k$ \\
    $\mathcal{R}^k$ & Redundant optimization component of client $k$ based on cached knowledge \\
    $\mathcal{L}_{KL}$ & Kullback-Leibler divergence loss \\
    $({zr}_{i}^{k})_s$ & The $s$-th knowledge fetched from the knowledge cache for sample index $(k,i)$ \\
    $\mathcal{D}^k_{distill}$ &  Distilled data on client $k$ \\
    $Sub^k$ & Adaptive sample strategy tailored for client $k$ \\
    $F^k_f$ & Feature extractor of the model on client $k$ \\
    $F^k_c$ & Classifier of the model on client $k$ \\
    $\sigma$ & Periodically updated random replacement function \\
    $\mathcal{D}^k_0$ & Subset of $\mathcal{D}^k$ with $C$ elements \\
    $\mathcal{L}^k_b$ & Dataset distillation objective on client $k$ \\
    $\mathcal{L}^k_{train}$ & Local training loss function on client $k$ \\
    $p^k_c$ & Label frequency of class $c$ on client $k$ \\
        \bottomrule
    \end{tabular}
    \label{notations-table}
\end{table}

\begin{equation}
    \mathop {\min }\limits_{\mathop  \cup \limits_{k = 1}^K \{ {W^k}\} }\frac{1}{K}\mathop \sum\limits_{k = 1}^K \left( {\frac{1}{{|{\mathcal{D}^k}|}}\sum\limits_{({X_i^k},{y_i^k}) \in {\mathcal{D}^k}} \mathcal{L}^i ({W^k};{X_i^k},{y_i^k})} \right).
\end{equation}
%Given the instability of device connections in edge environments, operations that require multiple clients to be online simultaneously (which we call synchronization) are generally discouraged. Ideally, the total number of synchronization operations across all $T$ communication rounds should be zero, which means:
Given the instability of device connections in edge environments, multiple clients may not be online simultaneously. 
%For any two clients $l$ and $m$, let $s_{lm}^t=0$ denote clients $l$ and $m$ are not online simultaneously at the $t$-th communication round, vice versa.}
% \begin{equation}
%     \sum\limits_{t = 0}^{T-1} {\sum\limits_{l = 1}^K {\sum\limits_{m = 1,l \ne m}^K {s_{lm}^t} } }  = 0,
% \end{equation}
% where $s_{lm}^t$ denotes the number of synchronization operations between clients $l$ and $m$ in the $t$-th communication round.
Besides, it is essential to minimize the communication overhead between devices and the server under the premise of guaranteeing user model accuracy \cite{mills2021multi}, saving valuable wireless network resources as well as device energy.

% while ensuring the accuracy of user models \cite{mills2021multi}. This approach can significantly conserve valuable wireless network resources and device energy.

% At the same time, it is essential to reduce the communication burden between devices and the server under the premise of guaranteeing user model accuracy \cite{mills2021multi}, which can effectively 

% Moreover, considering communication bandwidth between devices and the server, it is essential to reduce the bandwidth usage during FL training, while ensuring that the average  across clients meets a specified threshold $\varepsilon$, that is:
% \begin{equation}
% \color{red}
%     \begin{array}{l}
% \mathop {\min }\limits_{Comm} \quad Comm\\
% {\rm s.t.}\quad \sum\limits_{k = 1}^K {\frac{1}{K}UA({W^k};\bigcup\limits_{i = 1}^{|{\mathcal{D}^k}|} {X_i^k} ,\bigcup\limits_{i = 1}^{|{\mathcal{D}^k}|} {y_i^k} ) \ge } \varepsilon, 
% \end{array}
% \color{black}
% \end{equation}
% where ${UA({W^k};\bigcup\limits_{i = 1}^{|{\mathcal{D}^k}|} {X_i^k} ,\bigcup\limits_{i = 1}^{|{\mathcal{D}^k}|} {y_i^k} )}$ is the evaluated user model accuracy on client $k$.

%\noindent
\subsection{Knowledge Cache-driven Federated Learning}
We formulate knowledge cache-driven FL as a distributed optimization problem with the assistance of the remote knowledge cache $KC$ on the server, that is:
\begin{equation}
\begin{array}{l}
\mathop {\min }\limits_{\mathop  \cup \limits_{k = 1}^K \{ {W^k}\} } \frac{1}{K}\mathop \sum\limits_{k = 1}^K ( {\frac{1}{{|{\mathcal{D}^k}|}}(\sum\limits_{(X_i^k,y_i^k) \in {\mathcal{D}^k}} {{\mathcal{L}_{CE}}} ({W^k};\varphi ({F^k}(X_i^k)),y_i^k)} \\
\;\;\;\; \;\;\;\; \;\;\;\; + \beta \cdot \mathcal{R}^k({W^k};{KC})) ),
\end{array}
\end{equation}
where $\mathcal{L}_{CE}$ is the cross-entropy loss, $\varphi$ is the softmax function, $F^k$ is the prediction function of the model on client $k$. $\mathcal{R}^k$ represents the redundant optimization component of client $k$ based on cached knowledge, with corresponding weighting term $\beta$. As an example, FedCache \cite{wang2021knowledge} considers model outputs (logits) as knowledge, and optimizes local models based on cached related knowledge, that is:
\begin{equation}
    \begin{array}{l}
         \;\;\;\; {\mathcal{R}^k}  \\
          = \sum\limits_{(X_i^k,y_i^k) \in {\mathcal{D}^k}} {\mathcal{L}_{KL}(\varphi ({F^k}(X_i^k))||\varphi (\frac{1}{R}\sum\limits_{{{(zr_i^k)}_s} \in {KC}[k,i]} {{{(zr_i^k)}_s}} ))}, 
    \end{array}
\end{equation}
where $\mathcal{L}_{KL}$ is the Kullback-Leibler Divergence loss, ${(zr_i^k)}_s$ is the $s$-th knowledge fetched from the knowledge cache for sample index $(k,i)$, $R$ is a hyper-parameter that controls the number of related knowledge in FedCache. However, FedCache exhibits severe limitations in providing rich, distribution-aware information for personalized optimization over devices. The amount of information attainable from the remote knowledge cache is significantly restricted due to the design of small-scale logits interactions, as shown in Figure \ref{limitation} (a). This design fails to offer sufficient optimization information for clients, leading to performance bottlenecks of FedCache, as shown in Figure \ref{limitation} (c). Additionally, FedCache relies on task-specific data encoders to capture private sample relations, which restricts its applicability across varied data modalities and application tasks.

\begin{figure*}[t]
	\centering	\includegraphics[width=1.00\textwidth]{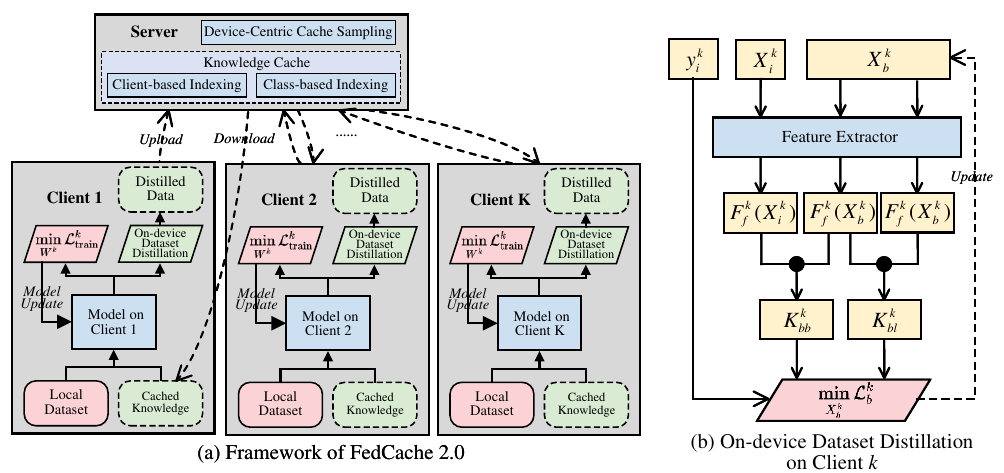}	\caption{Overview of FedCache 2.0.}
    \label{framework}
\end{figure*}

\begin{table*}[t]
\caption{\textcolor{black}{Comparison of FedCache 2.0 with state-of-art methods. Algorithms that achieve more than an order of magnitude reduction in communication overhead compared to model aggregation-based FL methods are classified as communication-efficient. User model accuracy is reported based on the experimental results described in Section \ref{exp-section}.}}
\label{compare}
\setlength{\tabcolsep}{8pt} 
\centering
\renewcommand\arraystretch{1.2}
\begin{tabular}{l|c|c|c|c|c}
\hline
\multicolumn{1}{c|}{\textbf{Method}} & \textbf{\begin{tabular}[c]{@{}c@{}}System\\ Heterogeneity\end{tabular}} & \textbf{\begin{tabular}[c]{@{}c@{}}Optimization\\ Personalization\end{tabular}} & \textbf{\begin{tabular}[c]{@{}c@{}}Efficient\\ Communication\end{tabular}} & \textbf{\begin{tabular}[c]{@{}c@{}}Uncertain Connection\\ Tolerance\end{tabular}} & \textbf{\begin{tabular}[c]{@{}c@{}}User Model\\ Accuracy\end{tabular}} \\ \hline
MTFL \cite{mills2021multi}                                 & \XSolidBrush                                                                       & \Checkmark                                                                               & \XSolidBrush                                                                          & \XSolidBrush                                                                                 & Low                                                                    \\
kNN-Per \cite{marfoq2022personalized}                             & \XSolidBrush                                                                       & \Checkmark                                                                               & \XSolidBrush                                                                          & \XSolidBrush                                                                                 & Medium                                                                 \\
SCDPFL \cite{chen2024spectral}                               & \XSolidBrush                                                                       & \Checkmark                                                                               & \XSolidBrush                                                                          & \XSolidBrush                                                                                 & High                                                                   \\
FedKD \cite{wu2022communication}                                & \Checkmark                                                                       & \XSolidBrush                                                                               & \Checkmark                                                                          & \XSolidBrush                                                                                 & Medium                                                                 \\
FedCache \cite{wu2024fedcache}                             & \Checkmark                                                                       & \Checkmark                                                                               & \Checkmark                                                                          & \Checkmark                                                                                 & High                                                                   \\ \hline
\textbf{FedCache 2.0}                & \textbf{\Checkmark}                                                              & \textbf{\Checkmark}                                                                      & \textbf{\Checkmark}                                                                 & \textbf{\Checkmark}                                                                        & \textbf{Highest}                                                       \\ \hline
\end{tabular}
\end{table*}
%\textbf{Personalized Optimization Reformulation.}
%\noindent
\subsection{FedCache 2.0 Optimization Formulation}
To address the aforementioned shortcomings of FedCache, FedCache 2.0 redesigns the original framework of FedCache by incorporating the benefits from both dataset distillation \cite{lei2023comprehensive}, offering a new interaction paradigm between devices and the server. \textcolor{black}{An overview of the FedCache 2.0 architecture is presented in Figure \ref{limitation}(b), and a detailed comparison with state-of-the-art methods is shown in Table \ref{compare}.} Specifically, local model optimization in FedCache 2.0 is regulated by post-sampled distilled data jointly synthesized by clients, that is:
\begin{equation}
{\mathcal{R}^k} = \sum\limits_{({X^*},{y^*}) \in {Sub}^k(\bigcup\limits_{l = 1}^K{\hat{\mathcal{D}}_{{distill}}^l})}  {{\mathcal{L}_{CE}}({W^k};\varphi ({F^k}({X^*})),{y^*})} ,
\end{equation}
where $\hat{\mathcal{D}}^l_{distill}$ is the synthetic data distilled on client $l$, $Sub^k$ represents the adaptive sample strategy tailered for client $k$. In our design, the synthesized data after sampling serves as the knowledge that devices request from the knowledge cache. This reformulation not only provides more comprehensive semantic information for local training on clients but also enhances the control over downloaded cached knowledge, enabling task-compatible and communication-efficient personalized optimization.

\section{FedCache 2.0}
In this section, we introduce our proposed FedCache 2.0 with an overview illustrated in Figure \ref{framework}. An execution procedure of FedCache 2.0 is elaborated in Algorithm \ref{alg}.

\subsection{Knowledge Cache Design}
Building upon the principles of knowledge-driven FL, FedCache 2.0 caches the latest distilled data as knowledge on the server side. In terms of knowledge cache operations, we provide two operations for indexing knowledge in the cache.

\noindent
\textbf{Client-Based Indexing.} Each client's distilled data is indexed by their identifier, allowing for efficient updates of knowledge in the cache and prototype initialization for on-device distillation, that is:
\begin{equation}
    KC[client,k] \leftarrow\hat{\mathcal{D}}^k_{distill},\forall k \in \{1,2,...,K\},
\end{equation}
where $KC$ is the notation of the knowledge cache on the server.

\noindent
\textbf{Class-Based Indexing.} All cached knowledge belonging to any specific class $y^*\in \{1,2,...,C\}$ are jointly fetched, facilitating the subsequent device-centric client sampling process, that is:
\begin{equation}
    S_c\leftarrow KC[class,c],\forall c \in \{1,2,...,C\},
\end{equation}
where $S_c$ the set of all knowledge belong to class $c$ in the knowledge cache, subject to:
\begin{equation}
\begin{array}{l}
    S_c=\{ ({X^*},{y^*})|({X^*},{y^*}) \in {KC}[{{client,k}}], \\
    k \in \{ 1,2,...,K\} ,{y^*} = c\}.
\end{array}
\end{equation}

\subsection{Federated Dataset Distillation}
FedCache 2.0 introduces federated dataset distillation, which collaboratively extracts anonymous structured information from local data on individual clients. This distilled data is stored on the server for further organization and accessibility.

\begin{algorithm*}[t]
\begin{spacing}{1.1}
\caption{FedCache 2.0.}
\label{alg}
\begin{minipage}{0.48\textwidth}
\begin{algorithmic}[1]
\State \textbf{procedure} ServerExecute()
\State \quad // Initialization Process
\State \quad \textbf{foreach} client $k \in \{1,2,\dots,K\}$:
\State \quad \quad $KC[client,k] \leftarrow \phi$
\State \quad \quad \textbf{foreach} class $c \in \{1,2,\dots,C\}$:
\State \quad \quad \quad Receive $p_c^k$ from client $k$
\State \quad // Training Process
\State \quad \textbf{foreach} client $k \in \{1,2,\dots,K\}$:
\State \quad \quad Send possible $\hat{\mathcal{D}}^k_b$ following Eq. (\ref{replacement})
\State \quad \quad Receive distilled data $\hat{\mathcal{D}}^k_b$ from client $k$
\State \quad \quad Update $KC$ following Eq. (\ref{update-kc})
\State \quad \quad Sample cache following Eq. (\ref{sample})
\State \quad \quad Send sampled knowledge to client $k$ 
\State \textbf{end procedure}
\end{algorithmic}
\end{minipage}%
\begin{minipage}{0.55\textwidth}
\begin{algorithmic}[1]
\State \textbf{procedure} ClientExecute($k$)
\State \quad // Initialization Process
\State \quad \textbf{foreach} class $c \in \{1,2,\dots,C\}$:
\State \quad \quad Compute $p_c^k$ following Eq. (\ref{p-k})
\State \quad \quad Send $p_c^k$ to the server
\State \quad // Training Process
\State \quad Initialize $\hat{\mathcal{D}}^k_b$ following Eq. (\ref{replacement})
\State \quad Compute $K_{bl}^k$ following Eq. (\ref{gram-1})
\State \quad Compute $K^k_{bb}$ following Eq. (\ref{gram-2})
\State \quad Optimize $\mathcal{L}_b^k$ following Eq. (\ref{dataset-distillation})
\State \quad Upload distilled data $\hat{\mathcal{D}}^k_b$ to server
%\State \quad // Collaborative Training
\State \quad Receive sampled knowledge from server
\State \quad Optimize $\mathcal{L}^k_{train}$ following Eqs. (\ref{training-loss},\ref{gating-function})
\State \textbf{end procedure}
\end{algorithmic}
\end{minipage}
\end{spacing}
\end{algorithm*}

\noindent
\textbf{On-Device Dataset Distillation.} All devices decompose their local models into feature extractors and classifiers. For each given sample $(X^*,y^*)\in \mathcal{D}^k$ on device $k$, the outputs of corresponding feature extractors and classifiers are denoted as $F^k_f(X^*)$ and $F^k_c(F^k_f(X^*))$, respectively. 
To start dataset distillation, each device $k$ initializes its prototype by selecting one local sample per class during the first communication round or receiving distilled data from other clients during subsequent communication rounds. The latter process is controlled by a periodically updated random replacement function $\sigma: \{1,2,...,K\} \rightarrow \{1,2,...,K\}$, with the intermediate distilled data stored in the knowledge cache, that is:
\begin{equation}
{\hat{\mathcal{D}}^k_b} \leftarrow \left\{ 
\begin{array}{l}
KC[client,\sigma (k)], \quad KC[client,\sigma (k)] \ne \phi \\ \quad\quad
{\mathcal{D}^k_0},\quad\quad\quad  \quad\quad KC[client,\sigma (k)] = \phi 
\end{array} \right.
\label{replacement},
\end{equation}
where ${\hat{\mathcal{D}}^k_b}$ denotes the set of prototype samples to be optimized into synthetic data after distillation. ${\mathcal{D}^k_0}$ is a subset of ${\mathcal{D}^k}$ with $C$ elements, subject to:
\begin{equation}
\begin{array}{l}
    y_0^k \ne y{_0^{k\prime} } \vee y_0^k = y{_0^{k\prime} } \wedge X_0^k = X{_0^{k\prime} }, \\
    \forall (X_0^k,y_0^k) \in \mathcal{D}_0^k \wedge (X{_0^{k\prime} },y{_0^{k\prime} }) \in \mathcal{D}_0^k.
\end{array}
\end{equation}
Without loss of generality, we assume device $k$ sets up a prototype $(X^k_{b},y^k_{b})\in {\hat{\mathcal{D}}^k_b}$ on class $y_{b}^k$. The on-device dataset distillation process should include computing the distance between the prototype's feature maps and those of the local data using the Gram matrix, that is:
\begin{equation}
    K_{bl}^k = F_f^k(X_i^k) \cdot F_f^k{(X_b^k)^{\rm T}}.
    \label{gram-1}
\end{equation}
Similarly, we compute the Gram matrix of the prototype itself:
\begin{equation}
    K^k_{bb}=F_f^k{(X_b^k)}\cdot F_f^k{(X_b^k)^{\rm T}}.
    \label{gram-2}
\end{equation}
The dataset distillation objective $\mathcal{L}_b^k$ is then optimized following kernel ridge regression loss:
\begin{equation}
    \mathop {\min }\limits_{X_b^k} \mathcal{L}_b^k = \mathop {\min }\limits_{X_b^k} \frac{1}{2}||y_b^k - K_{bl}^k{(K_{bb}^k + \lambda I)^{ - 1}} \cdot y_i^k|{|^2},
    \label{dataset-distillation}
\end{equation}
where $I$ denotes the identity matrix, and $\lambda$ is a hyper-parameter to control the degree of regularization. Note that local data is often augmented using common dataset enhancement techniques to increase the diversity of local feature maps during distillation. After obtaining the distilled data on client $k$, it is stored in the knowledge cache $KC$, ensuring the devices always have access to the latest distilled knowledge in the following communication rounds, that is:
\begin{equation}
    {KC[client,k]}\leftarrow {\hat{\mathcal{D}}^k_b}.
    \label{update-kc}
\end{equation}
In Appendix \ref{app-privacy}, we will further demonstrate the privacy guarantee of distilled data transmission from devices to the server.

\noindent
\textbf{Collaborative Training.} On-device dataset distillation relies on well-optimized feature extractors. To enhance local model performance and improve future distillation quality, devices periodically request cached distilled data from the server for personalized optimization. This collaborative training procedure is formulated as follows:

\begin{equation}
    \begin{array}{l}
    \;\;\;\;\min \limits_{{W^k}} \mathcal{L}^k_{train}
    \\=\mathop {\min }\limits_{{W^k}} \sum\limits_{(X_i^k,y_i^k) \in {\mathcal{D}^k}} {{\mathcal{L}_{CE}}({W^k};\varphi ({F^k}(X_i^k)),y_i^k)} \\
    \;\;\;\;+g(  \sum\limits_{({X^*},{y^*}) \in Su{b^k}(\bigcup\limits_{l = 1}^L {KC[client,l]} )} {{\mathcal{L}_{CE}}({W^k};\varphi ({F^k}(X^*)),y^*))} ,
    \end{array}
    \label{training-loss}
\end{equation}
where $\mathcal{L}^k_{train}$ denotes the local training loss function on client $k$, $g$ is a gating function acting as an identity mapping when the knowledge cache is empty in the first communication round and resulting in $0$ otherwise, that is:
\begin{equation}
    g(x) = \left\{ \begin{array}{l}x, \quad {KC[client,k]} \ne \phi \\0, \quad {KC[client,k]} = \phi\end{array} \right.,\forall x.
    \label{gating-function}
\end{equation}

\begin{figure}[t]
    \centering
    \begin{minipage}{\linewidth}
        \centering
        \includegraphics[width=1.0\textwidth]{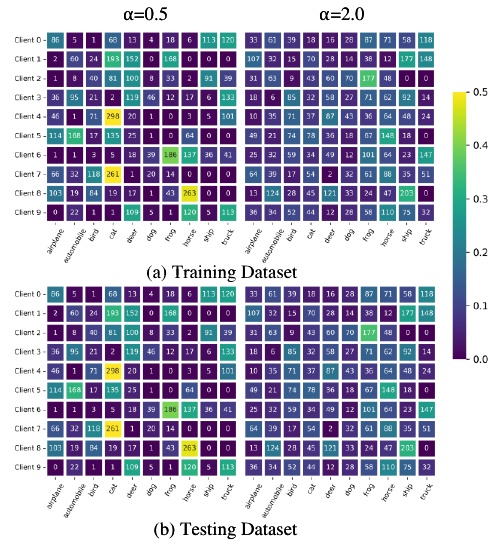}
        \caption{Illustration of varying degrees of data heterogeneity with different $\alpha$ across 10 clients over CIFAR-10 dataset. Each cell's color represents the proportion of samples in their respective datasets.}
        \label{appendix-data-dist}
    \end{minipage}\par\medskip
    
\end{figure}

\begin{table}[t]
\caption{Default hyper-parameters used in our experiments.}
\label{appendix-hypersetting}
\centering
\begin{adjustbox}{width=0.48\textwidth}
\renewcommand{\arraystretch}{1.0}
\setlength{\tabcolsep}{2pt} 
\begin{tabular}{c|c|c|c}
\hline
\textbf{Method}               & \textbf{Source of Code}                 & \textbf{Hyper-Parameters}  & \textbf{Value} \\ \hline
\multirow{10}{*}{MTFL}        & \multirow{10}{*}{\cite{mtflurl}}    & optimizer                  & customized     \\
                              &                                      & batch size                 & 64             \\
                              &                                      & learning rate              & 0.01           \\
                              &                                      & local epoch                & 1              \\
                              &                                      & communicaion round         & 100            \\
                              &                                      & server\_lr                 & 0.01           \\
                              &                                      & bn\_private                & usyb           \\
                              &                                      & $\beta_1$                      & 0.9            \\
                              &                                      & $\beta_2$                      & 0.999          \\
                              &                                      & $C$                          & 0.5            \\ \hline
\multirow{7}{*}{kNN-Per}      & \multirow{7}{*}{{}\cite{knnperurl}{}}  & optimizer                  & Adam           \\
                              &                                      & batch size                 & 64             \\
                              &                                      & learning rate              & 0.01           \\
                              &                                      & local epoch                & 1              \\
                              &                                      & communicaion round         & 100            \\
                              &                                      & aggregator\_type           & centralized    \\
                              &                                      & client\_type               & KNNPerClient   \\ \hline
\multirow{7}{*}{SCDPFL}       & \multirow{7}{*}{{}\cite{scdpflurl}{}}        & optimizer                  & Adam           \\
                              &                                      & batch size                 & 64             \\
                              &                                      & learning rate              & 0.01           \\
                              &                                      & local epoch                & 2              \\
                              &                                      & communicaion round         & 100            \\
                              &                                      & $\lambda_l$                  & 0.4            \\
                              &                                      & $\lambda_g$                  & 0.3            \\ \hline
\multirow{7}{*}{FedKD}        & \multirow{7}{*}{{}\cite{fedkdurl}{}}    & optimizer                  & Adam           \\
                              &                                      & batch size                 & 64             \\
                              &                                      & learning rate              & 0.01           \\
                              &                                      & local epoch                & 1              \\
                              &                                      & communicaion round         & 100            \\
                              &                                      & tmax                       & 0.98           \\
                              &                                      & tmin                       & 0.95           \\ \hline
\multirow{6}{*}{FedCache}     & \multirow{6}{*}{{}\cite{fedcacheurl}{}} & optimizer                  & Adam           \\
                              &                                      & batch size                 & 64             \\
                              &                                      & learning rate              & 0.01           \\
                              &                                      & local epoch                & 1              \\
                              &                                      & communicaion round         & 100            \\
                              &                                      & $\beta$                          & 1.5             \\
                              &                                      & $R$                          & 16             \\ \hline
\multirow{6}{*}{FedCache 2.0} & \multirow{6}{*}{Our Implementation}  & optimizer                  & Adam           \\
                              &                                      & batch size                 & 64             \\
                              &                                      & learning rate              & 0.01           \\
                              &                                      & local epoch                & 5              \\
                              &                                      & communicaion round         & 15             \\
                              &                                      & distillation learning rate & 0.001          \\
                              &                                      & $\tau$ & 0.5          \\\hline
\end{tabular}
\end{adjustbox}
\end{table}
\subsection{Device-Centric Cache Sampling}
To enhance personalized performance while reducing communication overhead, we propose a device-centric cache sampling strategy that considers local data characteristics and communication budgets.

\noindent
\textbf{Local Label Distribution Computation.} During the initialization process, each client $k$ computes its local label distribution according to its label frequency, that is:
\begin{equation}
    p_c^k = \frac{{|\{ (X_i^k,y_i^k)|(X_i^k,y_i^k) \in {\mathcal{D}^k},y_i^k = c\} |}}{{|{\mathcal{D}^k}|}},
    \label{p-k}
\end{equation}
where $p_c^k$ represents the label frequency of class $c$ on client $k$.

\noindent
\textbf{Distribution-Aware Controllable Sampling.} During the training process, the knowledge cache samples and distributes its stored knowledge based on $p_c^k$, that is:
\begin{equation}
\begin{array}{l}
    \;\;\;\; {Su{b^k}(\bigcup\limits_{l = 1}^L {KC[client,l]} )} \\
    {= \bigcup\limits_{c = 1}^C {RS(KC[class,c],(\tau  + (1 - \tau ) \cdot p_c^k) \cdot |KC[class,c]|)} },
    \label{sample}
\end{array}
\end{equation}
where $RS({\hat{\mathcal{D}}^*},p_0)$ denotes random sampling in the cached knowledge set ${\hat{\mathcal{D}}}^*$ at a probability $p_0$.
$\tau$ is a hyper-parameter ranging from 0 to 1 to control the trade-off between model performance and communication. As $\tau$ grows, the proportion of cached samples increases as well, leading to more cached knowledge but higher communication overhead.

% \vspace{10pt}
% \begin{figure}[t]
% 	\centering
% 	\includegraphics[width=1.0\textwidth]{communication-hetero.pdf}
% 	\caption{Average UA per unit of communication cost in experiments with heterogeneous models.}
%         \label{communication-hetero}
% \end{figure}

\begin{table*}[t]
\caption{Average UA (\%) on image recognition tasks with two degrees of data heterogeneity.}
\label{main-exp}
\setlength{\tabcolsep}{10pt} 
\centering
\renewcommand\arraystretch{1.2}
\begin{tabular}{c|l|cc|cc|cc}
\hline
\multirow{8}{*}{\textbf{\begin{tabular}[c]{@{}c@{}}Model\\ Homo.\end{tabular}}}   & \multicolumn{1}{c|}{\multirow{2}{*}{\textbf{Method}}} & \multicolumn{2}{c|}{\textbf{CIFAR-10}} & \multicolumn{2}{c|}{\textbf{CINIC-10}} & \multicolumn{2}{c}{\textbf{CIFAR-100}} \\
                                                                                  & \multicolumn{1}{c|}{}                                 & $\alpha$=0.5     & {$\alpha$=2.0}    & $\alpha$=0.5     & $\alpha$=2.0    & $\alpha$=0.5     & $\alpha$=2.0    \\ \cline{2-8} 
                                                                                  & MTFL                                                  & 31.1              & 29.2             & 32.1              & 34.8             & 14.8               & 15.4             \\
                                                                                  & kNN-Per                                               & 32.7              & 34.8             & 32.8              & 29.6             & 18.8              & 18.3             \\
                                                                                  & SCDPFL                                                & 49.4              & 33.1             & 48.7              & 32.2             & 33.3              & 19.6             \\
                                                                                  & FedKD                                                 & 40.9              & 23.9             & 39.2              & 22.7             & 26.1              & 14.3             \\
                                                                                  & FedCache                                              & 42.1              & 23.9             & 39.8              & 21.9             & 26.4              & 14.7             \\
                                                                                  & \textbf{FedCache 2.0}                                 & \textbf{51.1}     & \textbf{36.5}    & \textbf{51.1}     & \textbf{36.3}    & \textbf{35.8}     & \textbf{23.3}    \\ \hline
\multirow{5}{*}{\textbf{\begin{tabular}[c]{@{}c@{}}Model\\ Hetero.\end{tabular}}} & \multicolumn{1}{c|}{\multirow{2}{*}{\textbf{Method}}} & \multicolumn{2}{c|}{\textbf{CIFAR-10}} & \multicolumn{2}{c|}{\textbf{CINIC-10}} & \multicolumn{2}{c}{\textbf{CIFAR-100}} \\
                                                                                  & \multicolumn{1}{c|}{}                                 & $\alpha$=0.5     & $\alpha$=2.0    & $\alpha$=0.5     & $\alpha$=2.0    & $\alpha$=0.5     & $\alpha$=2.0    \\ \cline{2-8} 
                                                                                  & FedKD                                                 & 39.7              & 24.1             & 39.6              & 23.6             & 26.2              & 14.1             \\
                                                                                  & FedCache                                              & 41.3              & 22.2             & 40.3              & 22.4             & 26.3              & 13.9             \\
                                                                                  & \textbf{FedCache 2.0}                                 & \textbf{51.1}     & \textbf{35.7}    & \textbf{51.2}     & \textbf{36.9}    & \textbf{35.8}     & \textbf{23.5}    \\ \hline
\end{tabular}
%\end{table*}

\vspace{15pt}

%\begin{table*}[t]
\centering
    \caption{Communication cost and efficiency speed-up ratio in image recognition task under two degrees of data heterogeneity with experiments on homogeneous models. The communication cost is measured when the average UA reaches the given threshold. Some methods fail to achieve the average UA threshold, and their communication costs are denoted as N/A. The same as below.}
    \label{appendix-communication-image}
    \setlength{\tabcolsep}{10pt} 
%\begin{adjustbox}{width=0.47\textwidth}
\renewcommand\arraystretch{1.2}
\begin{tabular}{c|l|ccc}
\hline
\multirow{7}{*}{$\bm{\alpha}$\textbf{=0.5}} & \multicolumn{1}{c|}{\textbf{Method}} & \textbf{CIFAR10/45\%} & \textbf{CINIC10/40\%} & \textbf{CIFAR100/30\%} \\ \cline{2-5} 
                                & MTFL                                 & N/A                   & N/A                   & N/A                    \\
                                & kNN-Per                              & N/A                   & N/A                   & N/A                    \\
                                & SCDPFL                               & 17.4G ($\times$1.0)             & 7.8G ($\times$1.0)              & 17.6G ($\times$1.0)              \\
                                & FedKD                                & N/A                   & N/A                   & N/A                    \\
                                & FedCache                             & N/A                   & N/A                   & N/A                    \\
                                & \textbf{FedCache 2.0}                         & \textbf{389.1M ($\times$45.8)}         & \textbf{193.8M ($\times$41.2)}         & \textbf{609.3M ($\times$29.6)}          \\ \hline
\multirow{7}{*}{$\bm{\alpha}$\textbf{=2.0}} & \multicolumn{1}{c|}{\textbf{Method}} & \textbf{CIFAR10/30\%} & \textbf{CINIC10/30\%} & \textbf{CIFAR100/15\%} \\ \cline{2-5} 
                                & MTFL                                 & N/A                   & 39.9G ($\times$1.0)             & 64.5G ($\times$1.0)              \\
                                & kNN-Per                              & 12.1G ($\times$1.6)           & N/A                   & 17.6G ($\times$3.7)            \\
                                & SCDPFL                               & 19.4G ($\times$1.0)             & 10.6G ($\times$3.8)           & 15.9G ($\times$4.1)            \\
                                & FedKD                                & N/A                   & N/A                   & N/A                    \\
                                & FedCache                             & N/A                   & N/A                   & N/A                    \\
                                & \textbf{FedCache 2.0}                         & \textbf{315.9M ($\times$62.9)}         & \textbf{473.5M ($\times$86.3)}         & \textbf{197.5M ($\times$334.4)}         \\ \hline
\end{tabular}
%\end{adjustbox}
\end{table*}

\section{Experiments}
\label{exp-section}
\subsection{Experimental Setup}
\textbf{Platforms.} 
Our experiments are conducted on a high-performance physical server equipped with  12th Gen Intel(R) Core(TM) i7-12700 CPU and multiple NVIDIA GeForce RTX 3090 GPU cards. The server's memory consists of four 16GB Acer DDR4 modules operating at 2133 MT/s, providing a total of 64GB of RAM. Storage is handled by a KINGSTON SKC3000D2048G solid-state drive.

\noindent
\textbf{Datasets.} We evaluate the effectiveness of our proposed FedCache 2.0 across various application tasks, including image recognition, audio understanding, and mobile sensor data mining \cite{carpineti2018custom}. These experiments cover five datasets: which are CIFAR10, CIFAR100 \cite{cifar10}, CINIC10 \cite{cinic10}, UrbanSound8K \cite{urbansound}, and TMD \cite{carpineti2018custom}. Detailed illustrations of tasks and datasets are provided in Appendix \ref{appendix-dataset}. Each complete dataset is preprocessed using the distributed data partition strategy from FedML \cite{he2020fedml}, with a hyper-parameter $\alpha$ to adjust the degree of data heterogeneity among clients.

\noindent
\textbf{Models.} We employ five model structures, considering both deep residual network \cite{he2016deep} for image data, and fully connected network for numeric data. In addition, we consider both scenarios with homogeneous and heterogeneous on-device models in our experiments, with detailed model settings provided in section \ref{appendix-implementation-details}.

\noindent
\textbf{Baselines.} We compare FedCache 2.0 against the following state-of-the-art methods: MTFL \cite{mills2021multi}, KNN-Per \cite{marfoq2022personalized}, spectral co-distillation for personalized FL (SCDPFL) \cite{chen2024spectral}, FedKD \cite{wu2022communication} and FedCache \cite{wu2024fedcache}. These baseline algorithms encompass personalized/multi-task FL methods, FL algorithms addressing dual model heterogeneity and communication efficiency, and FL for edge computing. 
%Implementation details for these algorithms are available in Appendix \ref{appendix-implementation-details}. 

\noindent
\textbf{Criteria.} Following \cite{mills2021multi}, we adopt the average User model Accuracy (UA) as the primary metric for evaluating model precision, focusing on the highest value achieved within 100 communication rounds. In addition, we assess communication efficiency by monitoring the learning curves, measuring average UA against per unit of communication overhead. Detailed elaborations on how communication cost is calculated are provided in Appendix \ref{appendix-communication-computation}.

\begin{figure*}[t]
	\centering
	\includegraphics[width=1.0\textwidth]{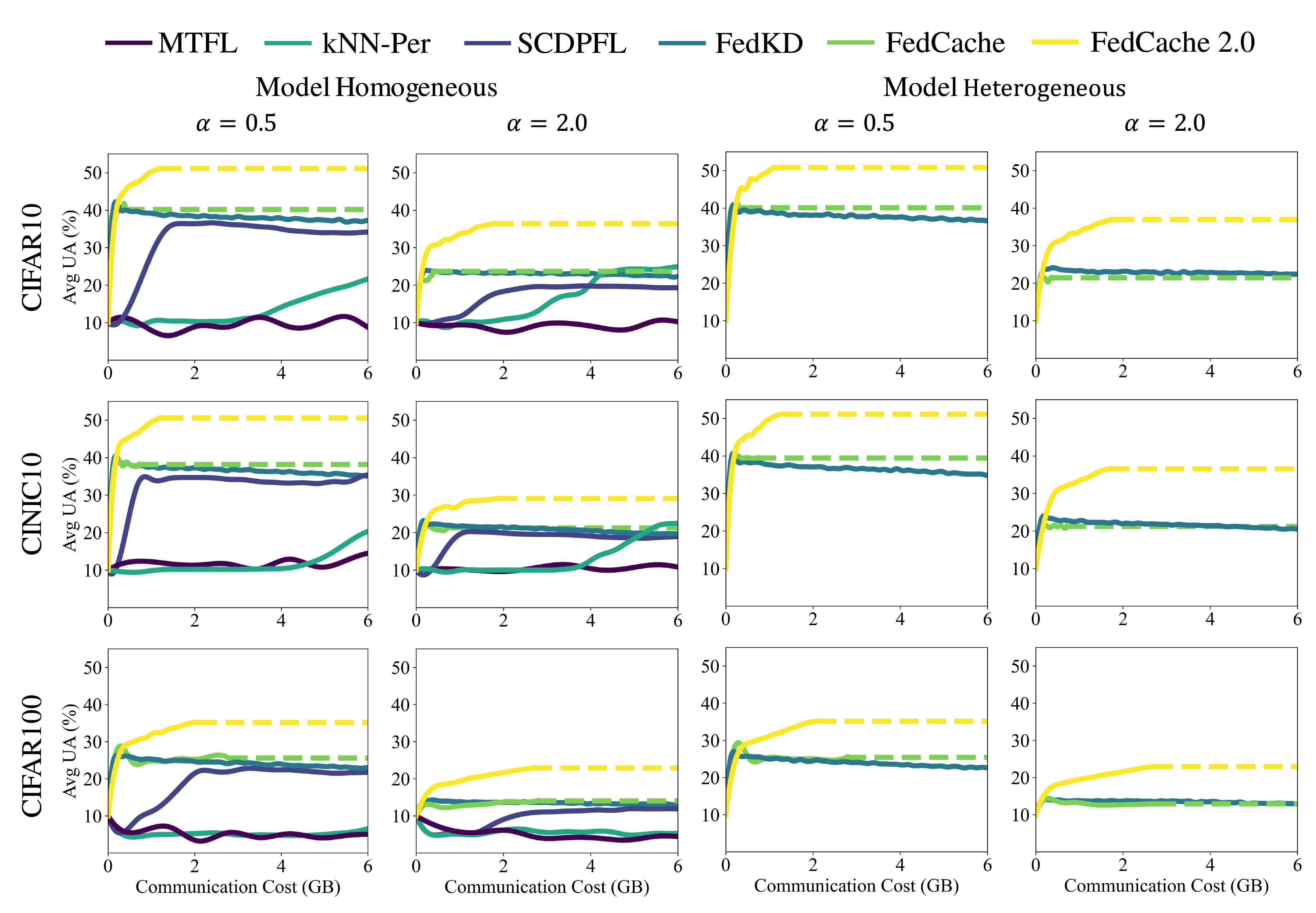}
	\caption{Average UA UA (\%) per unit of communication cost over image recognition tasks.}
        \label{communication}
\end{figure*}

\color{black}
\subsection{Implementation Details}
    \label{appendix-implementation-details}
    % In this section, we outline the implementation details of our experiments, elaborating on the model configurations for both homogeneous and heterogeneous settings, the adopted data partition strategy, and the hyper-parameters employed across our experiments. 

   \noindent
    \textbf{Data Partition.} For all experiments conducted in this paper, we utilize the data partition strategy provided by FedML \cite{he2020fedml} to simulate $K=100$ decentralized datasets for clients. We adopt the hyper-parameter $\alpha \in \{0.5, 2.0\}$ to control the degree of data heterogeneity. A smaller number of $\alpha$ results in a greater degree of heterogeneity among clients. Notably, the training and testing datasets for the same client have identical distributions, whereas different clients typically have different data distributions, as illustrated in Figure \ref{appendix-data-dist}.

    \noindent
    \textbf{Model Configurations.} We conduct experiments in both model-homogeneous and model-heterogeneous settings respectively, with detailed information on model structures provided in Appendix \ref{appendix-model-structure-configurations}.    
    For image recognition experiments with homogeneous models, all clients employ ResNet-L. For image recognition experiments with heterogeneous models, models on clients are evenly distributed among ResNet-S, ResNet-M, and ResNet-L. Note that for this set of experiments, we only consider FedKD \cite{wu2022communication} and FedCache \cite{wu2024fedcache} as baseline algorithms to compare with. This choice is based on the fact that MTFL \cite{mills2021multi}, kNN-Per \cite{marfoq2022personalized}, and SCDPFL \cite{chen2024spectral} support only homogeneous models across clients. In addition, ResNet-T is adopted to facilitate the interaction of model parameters between clients and the server within the FedKD algorithm in all image recognition experiments. For audio understanding on UrbanSound8K \cite{urbansound} and mobile sensor data mining on TMD \cite{carpineti2018custom}, we exclusively employ the model-homogeneous setting, with all clients deploying a uniform FCN model, whose structures are denoted as FCN-U and FCN-T, respectively.

    \noindent
    \textbf{Hyper-parameter Settings.} We provide a default setting in Table \ref{appendix-hypersetting}, which is implemented in all experiments unless stated otherwise. Note that the table only showcases a subset of hyperparameters for baseline algorithms; those not mentioned retain their original settings as specified in the corresponding open-source code. Exclusively, we uniformly set the learning rate to 0.1 and the communication round of kNN-Per to 200 for experiments on mobile sensor data mining.
\color{black}

\subsection{Evaluation on Image Classification}
\textbf{Average User Model Accuracy.} Table \ref{main-exp} displays the comparison of average UA on image recognition datasets, CIFAR10, CIFAR100, and CINIC10, with two degrees of data heterogeneity, $\alpha\in \{0.5,2.0\}$. As displayed, FedCache 2.0 significantly outperforms all considered state-of-the-art methods across both model homogeneous and heterogeneous settings, demonstrating its superior performance and robustness in diverse edge scenarios. This substantial improvement is attributed to the enriched information characterization provided by distilled data and effective personalized optimization facilitated by device-centric cache sampling. %Additional evaluations on audio understanding and mobile sensor data mining are provided in Appendix \ref{appendix-results-downstream-tasks}.

\begin{table}[t]
  \centering
  \begin{minipage}{0.5\textwidth}
    \centering
    \caption{Average UA UA (\%) on audio understanding with two degrees of data heterogeneity.}
    \label{appendix-urbansound}
    \renewcommand\arraystretch{1.2}
    \begin{tabular}{l|cc}
\hline
\multicolumn{1}{c|}{\textbf{Method}} & $\bm{\alpha=0.5}$ & $\bm{\alpha=2.0}$ \\ \hline
MTFL                                 & 56.5           & 46.9           \\
kNN-Per                              & 54.3           & 54.1           \\
\textbf{FedCache 2.0}                & \textbf{69.4}  & \textbf{64.8}  \\ \hline
\end{tabular}
    
%\end{table}

\vspace{15pt}
    
%\begin{table}[t]
\centering
\setlength{\tabcolsep}{5pt} 
    \caption{Communication cost and efficiency speed-up ratio in audio understanding task under two degrees of data heterogeneity.}
    \renewcommand\arraystretch{1.2}
    \label{appendix-communication-audio}
\begin{tabular}{l|cc}
\hline
\multicolumn{1}{c|}{\textbf{Method}} & \textbf{$\bm{\alpha=0.5}$/50\%}     & \textbf{$\bm{\alpha=2.0}$/45\%}     \\ \hline
MTFL                                 & 13.0G ($\times$1.0)            & 18.4G ($\times$1.0)            \\
kNN-Per                              & 7.5G ($\times$1.73)            & 7.2G ($\times$2.6)             \\
\textbf{FedCache 2.0}                & \textbf{14.3M ($\times$930.9)} & \textbf{19.1M ($\times$986.5)} \\ \hline
\end{tabular}

\vspace{13pt}
\includegraphics[width=\textwidth]{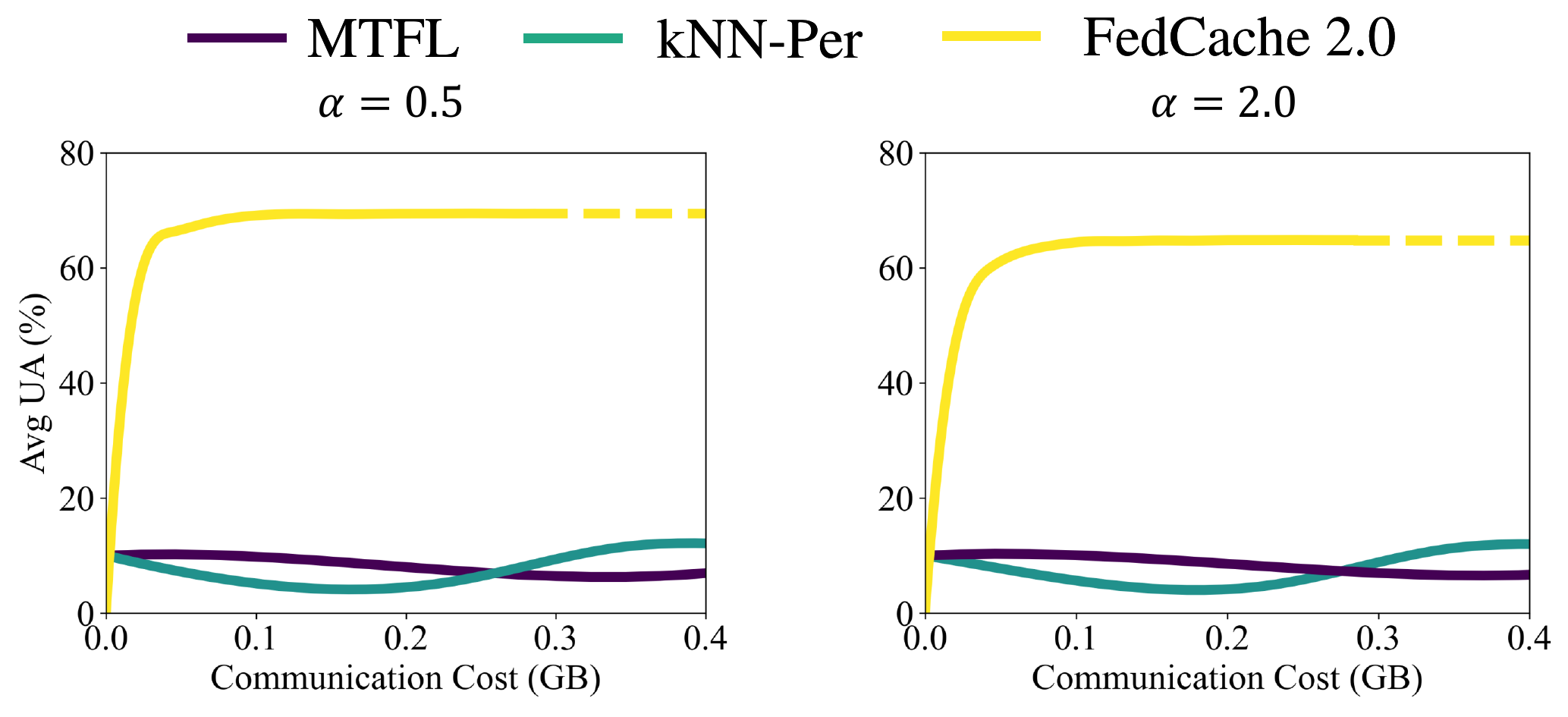}
        \caption{Average UA per unit of communication cost over audio understanding task.}
        \label{appendix-learning-audio}
        
\end{minipage}
\end{table}

\noindent
\textbf{Communication Cost.} Figure \ref{communication} illustrates the learning curves for image recognition tasks, plotting average UA against communication cost. As shown, FedCache 2.0 exhibits significantly steeper convergence curves, reaching acceptable average UA more efficiently than competing methods, regardless of data heterogeneity, model structures, and datasets. This indicates that FedCache 2.0 can achieve robust performance improvement with reduced communication overhead, making it suitable for deployment in resource-constrained edge environments with limited wireless bandwidth. The reduction in communication cost is attributed to FedCache 2.0's elimination of transferring cumbersome model parameters between devices and the server. Alternatively, FedCache 2.0 leverages compact distilled data as knowledge to facilitate communication-efficient personalized optimization on devices. \textcolor{black}{We also provide detailed quantitative communication burden for all considered FL algorithms in image recognition tasks with homogeneous models. We measure the communication cost in terms of the amount of information transmitted between clients and the server until the average User model Accuracy (UA) reaches a specified threshold. We also provide the communication efficiency speed-up ratio, which compares the communication cost of each method to that of the least efficient baseline method achieving the same threshold precision. Table \ref{appendix-communication-image} quantitates the communication cost and efficiency speed-up ratio of FedCache 2.0 compared with baseline algorithms. It is evident that FedCache 2.0 significantly reduces the communication cost to reach a given target average UA compared to baseline algorithms, with 28.6 to 332.4 times of communication efficiency improvements compared with state-of-the-art methods.} %Additional experimental results on audio understanding and mobile sensor data mining are provided in Appendix \ref{appendix-results-downstream-tasks}.

% \begin{figure}[!t]
%         \centering
        
% \end{figure}
%\vspace{10pt}

\begin{table}

  \hfill
  \begin{minipage}{0.5\textwidth}
    \centering
    \caption{Average UA on mobile sensor data mining with two degrees of data heterogeneity.}
    \label{appendix-tmd}
    \renewcommand\arraystretch{1.2}
    \begin{tabular}{l|cc}
\hline
\multicolumn{1}{c|}{\textbf{Method}} & $\bm{\alpha=0.5}$         & $\bm{\alpha=2.0}$         \\ \hline
MTFL                        & 54.8          & 46.5          \\
kNN-Per                     & 61.6          & 48.6          \\
\textbf{FedCache 2.0}       & \textbf{73.3} & \textbf{61.7} \\ \hline
\end{tabular}

  \end{minipage}
  \vspace{15pt}
%\begin{table}[t]

\centering
    \caption{Communication cost and efficiency speed-up ratio in mobile sensor data mining task under two degrees of data heterogeneity.}
    \label{appendix-communication-tmd}
    \renewcommand\arraystretch{1.2}
\begin{tabular}{l|cc}
\hline
\multicolumn{1}{c|}{\textbf{Method}} & \textbf{$\bm{\alpha=0.5}$/50\%}  & \textbf{$\bm{\alpha=2.0}$/45\%}   \\ \hline
MTFL                                 & 20.5G   ($\times$1.0)           & 11.9G   ($\times$1.0)            \\
kNN-Per                              & 12.4G   ($\times$1.7)           & 11.0G   ($\times$1.1)            \\
\textbf{FedCache 2.0}                & \textbf{14.8M ($\times$1418.4)} & \textbf{15.4M   ($\times$791.3)} \\ \hline
\end{tabular}

\vspace{13pt}

\includegraphics[width=0.5\textwidth]{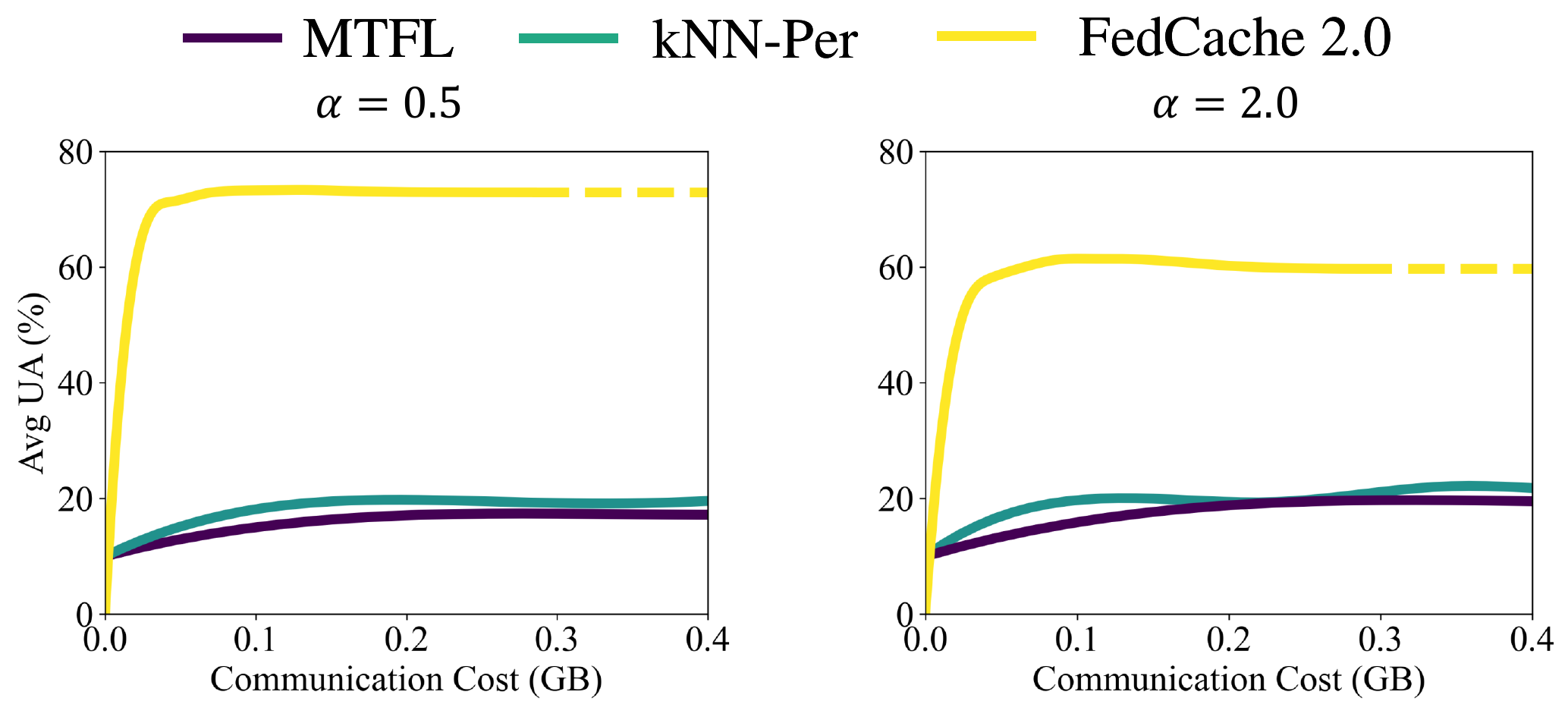}
        \caption{Average UA per unit of communication cost over mobile sensor data mining task.}
        \label{appendix-learning-tmd}
\end{table}

\color{black}
\subsection{Evaluation on Audio Understanding}
    Table \ref{appendix-urbansound} presents the average UA on UrbanSound8K \cite{urbansound} dataset under two degrees of data heterogeneity. As observed, FedCache 2.0 consistently outperforms the baseline methods, achieving higher UA across both settings. Additional experimental results on communication efficiency are provided in Table \ref{appendix-communication-audio} and Figure \ref{appendix-learning-audio}, demonstrating significant reductions in communication cost while maintaining superior performance.

    \subsection{Evaluation on Mobile Sensor Data Mining}
    Table \ref{appendix-tmd} presents the average UA on TMD \cite{carpineti2018custom} dataset under two degrees of data heterogeneity. Results demonstrate that FedCache 2.0 significantly outperforms considered baseline algorithms regardless of data heterogeneity settings. We also provide qualitative and quantitative evaluation of FedCache 2.0 in Figure \ref{appendix-learning-tmd} and Table \ref{appendix-communication-tmd}, respectively, indicating that FedCache 2.0 not only achieves higher system performance but also requires significantly less communication overhead.
\color{black}

% \begin{figure}[t]
% \centering
        
% \end{figure}

\subsection{Ablation Study}
\textbf{Impact of Cache Sampling Strategy.} 
Table \ref{ablation-tau} presents the average UA with different $\tau$ values. We can conclude that increasing $\tau$ in the early stage generally improves the average UA due to the richer information provided by a greater number of cached samples. However, the performance gain diminishes as $\tau$ approaches 1. This decline is likely due to the introduction of data distribution bias across devices, which is harmful to the system's performance.

\noindent
\textbf{Impact of Model Settings.} Table \ref{ablation-model} presents the average UA for different model configurations across image recognition datasets. The results indicate that heterogeneous model settings yield higher average UA compared to homogeneous settings constrained by the weakest end devices. The improvements stem from the support of more powerful devices to deploy larger and more complex models, which can make full use of computational resources among heterogeneous devices to achieve better performance. These findings underscore the benefits of model heterogeneity flexibility in FedCache 2.0.

\begin{table*}[t]
\caption{Ablation study on cache sampling strategy. Results are derived from the CIRAR-10 dataset with homogeneous models, taking $\alpha=0.5$.}
\label{ablation-tau}
\setlength{\tabcolsep}{10pt} 
\centering
\renewcommand\arraystretch{1.2}
\begin{tabular}{l|ccccc}
\hline
\multicolumn{1}{c|}{\textbf{Method}} & \textbf{$\bm{\tau=0}$} & \textbf{$\bm{\tau=0.3}$} & \textbf{$\bm{\tau=0.5}$} & \textbf{$\bm{\tau=0.7}$} & \textbf{$\bm{\tau=1.0}$} \\ \hline
FedCache                             & \multicolumn{5}{c}{41.2}                                                                  \\ \hline
\textbf{FedCache 2.0}                & 51.3          & \textbf{51.7}   & 51.1            & 49.8            & 48.6            \\ \hline
\end{tabular}
\end{table*}
\begin{table*}
\setlength{\tabcolsep}{10pt} 
%\begin{table}[]
\caption{Ablation study on model settings. Results are derived from image recognition tasks, taking $\alpha=0.5$.}
\label{ablation-model}
\centering
%\begin{adjustbox}{width=0.47\textwidth}
\renewcommand\arraystretch{1.2}
\begin{tabular}{l|c|ccc}
\hline
\multicolumn{1}{c|}{\textbf{Method}}   & \textbf{Model}             & \textbf{CIFAR10} & \textbf{CINIC10} & \textbf{CIFAR100} \\ \hline
\multirow{2}{*}{FedCache}              & ResNet-S                   & 40.5            & 38.3            & 25.2             \\
                                       & ResNet-S/ResNet-M/ResNet-L & 41.3            & 40.3            & 26.3             \\ \hline
\multirow{2}{*}{\textbf{FedCache 2.0}} & ResNet-S                   & 46.6            & 46.9             & 31.5             \\
                                       & ResNet-S/ResNet-M/ResNet-L & \textbf{51.1}   & \textbf{51.1}   & \textbf{35.8}    \\ \hline
\end{tabular}
%\end{adjustbox}
\end{table*}

\section{Discussion}
\subsection{Broader Impacts} 
FedCache 2.0 introduces tolerance for uncertain connection, allowing devices to engage in FEL at their convenience, without relying on any two devices being online at the same time. This feature is particularly advantageous in dynamic network environments, especially for IoT applications prone to unstable connectivity from power outages or limited signal coverage.
Moreover, FedCache 2.0's generalized data anonymization broadens its utility across various data modalities and application tasks. For instance, FedCache can be seamlessly integrated into smart healthcare and e-commerce recommendation systems, facilitating personalized trained models' deployment on smartwatches or mobile phones to monitor users' health status and shopping preferences.

% \noindent
\subsection{Limitations}
One of the primary concerns with FedCache 2.0 is the vulnerability to malicious participants. Since devices upload distilled data to the server, there is a risk that malicious devices could inject poisoned distilled data into the system. This could distort the knowledge cache and degrade the performance of the entire system. 
Another limitation of FedCache 2.0 is the computational burden imposed by the dataset distillation process on devices. Federated dataset distillation introduces additional optimization burdens besides local model training, demanding considerable computational resources. This request can somewhat lead to slower training procedures on devices with low hardware capabilities or those constrained by battery life.

%In terms of potential limitations of FedCache 2.0, devices may maliciously upload misleading or poisoned distilled data to the server, which could negatively affect the overall system performance. In addition, the dataset distillation process conducted on devices demands considerable computational resources. This request can somewhat lead to slower training procedures on devices with low hardware capabilities or those constrained by battery life.

\section{Related Work}
\subsection{Federated Learning with Dual Learning and Model Personalization}
\textcolor{black}{FL methods are designed with the motivation to adapt to various hardware configurations through model personalization, while also addressing diverse user preferences via learning personalization.
A variety of approaches have been developed to tackle the dual personalization challenges within FL.} Differentiated client-side model optimization objectives are implemented in studies such as \cite{marfoq2022personalized,shi2023prior,mills2021multi,jin2022personalized,spectral,fedfed}, enabling trained models to generalize across clients with varying local data distributions. Novel client-server interaction designs, which depart from the traditional FedAvg \cite{mcmahan2017communication}, are explored in \cite{wu2022communication,wang2023towards,zhu2021data,wu2024exploring,huang2022learn} to better accommodate diverse client hardware configurations with differently structured models. Furthermore, hybrid approaches such as \cite{wu2024fedcache,wu2023fedict} are proposed to simultaneously address both model and learning personalization in FL.

\textcolor{black}{Different from prior works, FedCache 2.0 implements a customized knowledge cache that stores and organizes distilled data to assist local model training, allowing heterogeneous clients to benefit from personalized knowledge fetched from the remote knowledge cache with efficient communication.}

%\noindent
\subsection{Federated Learning in Edge Computing} 
The efficiency of executing FL at the network edge has become a hot topic. Research works such as \cite{alam2022fedrolex,ijcai2022p399,he2020group,wen2024rtifed} investigate the technical frameworks required for running FL algorithms on devices constrained by computational power or storage resources. Device heterogeneity and connection uncertainty in edge environments are tackled by methodologies such as \cite{zhu2022resilient,wu2024fedcache,liu2023adaptive,wu2023fedict}. Moreover, \cite{wang2022accelerating,wu2023agglomerative,deng2023hierarchical,liu2020client,wang2021resource} extends FL to a multi-tier architecture involving end-edge-cloud collaborations, enhancing model training efficiency and final performance by leveraging the edge as a bridge between devices and the cloud during the training process.

\textcolor{black}{FedCache 2.0 is well-suited for edge environments as it effectively addresses key challenges such as limited bandwidth, device heterogeneity, and uncertain connectivity. By caching distilled data on the server and conducting device-centric cache sampling, FedCache 2.0 achieves state-of-the-art performance with reduced communication overhead, while ensuring that even resource-constrained devices can efficiently utilize global knowledge for locally adapted model training.}

%FedCache 2.0 takes a different approach by leveraging dataset distillation to transfer synthetic knowledge instead of large model parameters. This not only reduces communication costs but also allows for more flexible and asynchronous communication between clients and the server. By caching distilled data on the server, FedCache 2.0 avoids the need for clients to be online simultaneously, further enhancing its suitability for edge environments with intermittent connectivity.

%\noindent
\subsection{Federated Learning with Alternative Information} 
Instead of transmitting model parameters, alternative information is utilized in the FL training process by a series of recent works. Model-agnostic outputs are exchanged between clients and the server in \cite{wu2024fedcache,itahara2021distill,huang2022learn,wu2024exploring,li2019fedmd}, allowing deployment of customized on-device models across resource-heterogeneous clients. Additionally, methodologies involving uploading mixed or distilled data from clients to servers are proposed in \cite{song2023federated,oh2020mix2fld}, significantly reducing communication overhead while maintaining client data privacy.

\textcolor{black}{FedCache 2.0 extends these ideas by leveraging distilled data as the medium of knowledge exchange. Through implementing knowledge caching with federated dataset distillation, FedCache 2.0 provides a more performance-guaranteed paradigm for information transfer between clients and the server, while ensuring efficient communication.}

%FedCache 2.0 builds upon these ideas by combining knowledge caching with federated dataset distillation. By storing and transferring distilled synthetic data as knowledge, FedCache 2.0 offers a more performance-guaranteed paradigm to exchange information between clients and the server. 

%The use of distilled data allows FedCache 2.0 to capture essential information from client datasets while minimizing the communication burden, making it well-suited for real-world edge computing scenarios where bandwidth and computational resources are limited.

\section{Conclusion}
In this paper, we introduce FedCache 2.0, a novel personalized FEL architecture to address the challenges of resource heterogeneity, communication limitations, and dynamic network conditions in edge environments. By incorporating the benefits of both knowledge cache-driven federated learning and dataset distillation, FedCache 2.0 facilitates privacy-preserving and semantically enriched knowledge organization and transfer among devices and the server. This is achieved through an iterative process of distilling data on devices, caching them on the server, and then dispatching the cached knowledge to guide local training and subsequent distillation. Moreover, we propose a device-centric cache sampling strategy to further enhance personalized model training by adapting to client data distributions and communication constraints. Extensive experiments on various tasks and datasets demonstrate that FedCache 2.0 outperforms state-of-the-art methods with reduced communication costs, illustrating its potential as a promising solution for personalized edge intelligence scenarios.

\bibliographystyle{IEEEtran}

% \bibliographystyle{IEEEtran}
% \bibliography{ref.bib}
\vspace{-33pt}

\begin{IEEEbiography}[{\includegraphics[width=1in,height=1.25in,clip,keepaspectratio]{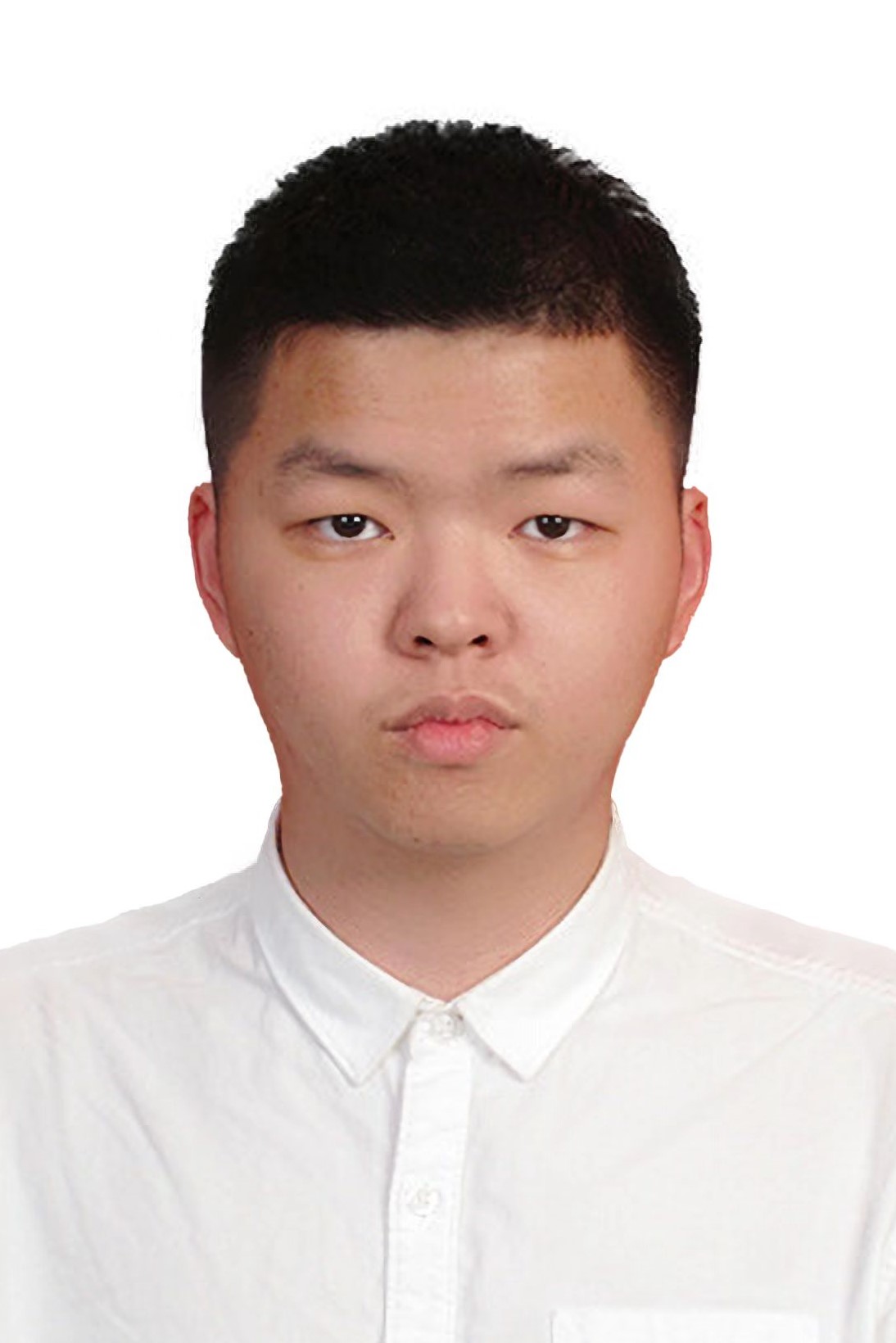}}]{Quyang Pan}
	is currently a research assistant with the Institute of Computing Technology, Chinese Academy of Sciences. He has published several technical papers at top-tier conferences and journals as the first author in the fields of computer architecture, computer networks, and intelligent systems, including IEEE Transactions on Parallel and Distributed Systems (TPDS), IEEE Transactions on Mobile Computing (TMC), IEEE International Conference on Computer Communications (INFOCOM), and ACM Transactions on Intelligent Systems and Technology (TIST). He is also an outstanding competitive programmer who has won several gold medals in international and national contests such as ACM International Collegiate Programming Contest (ICPC), CCF Collegiate Computer Systems and Programming Contest (CCSP), etc. His research interests include federated learning and edge computing.
\end{IEEEbiography}

\begin{IEEEbiography}[{\includegraphics[width=1in,height=1.25in,clip,keepaspectratio]{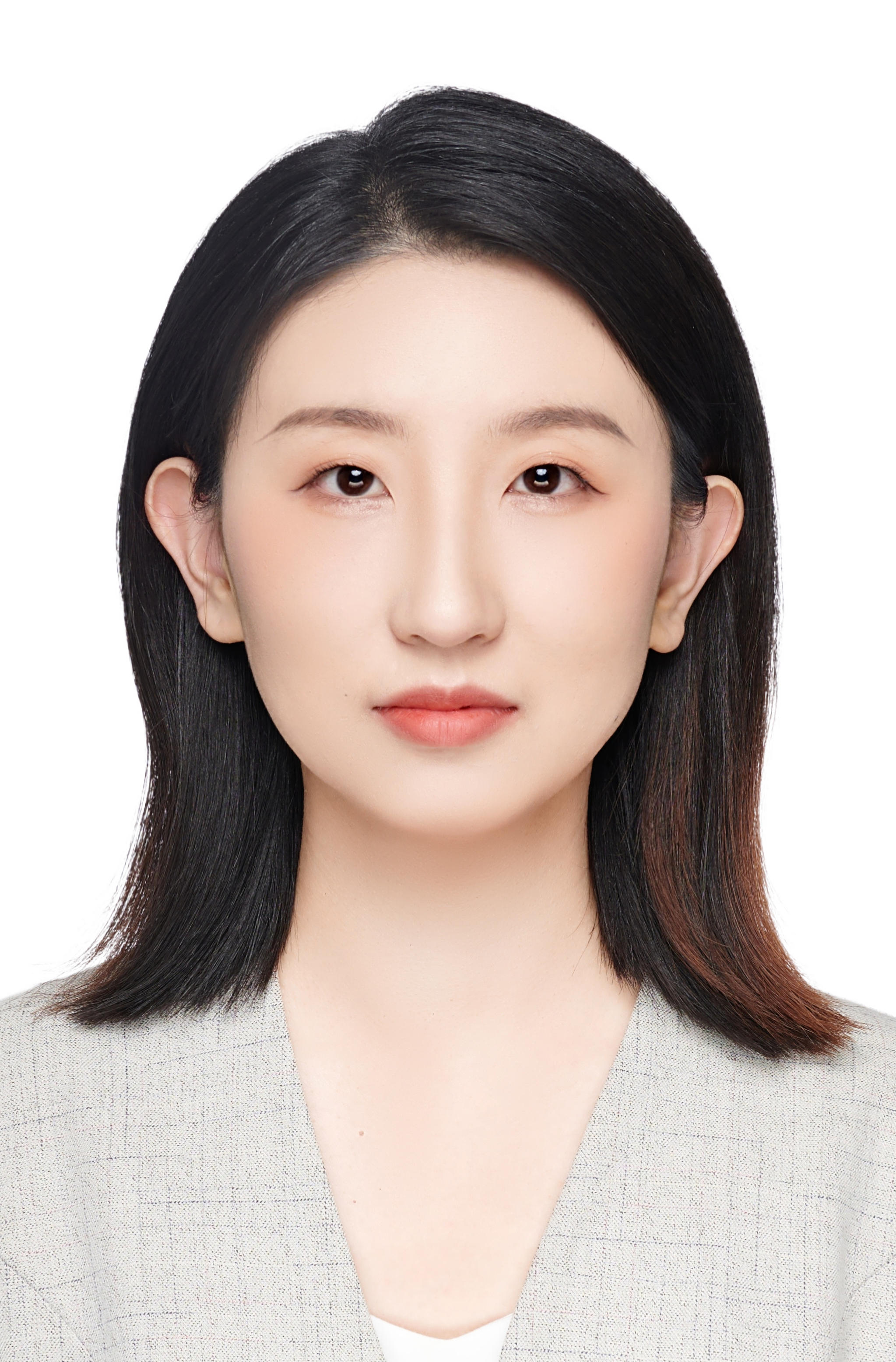}}]{Sheng Sun} is currently an associate professor at the Institute of Computing Technology, Chinese Academy of Sciences. She received her bachelor's degree from Beihang University, and her Ph.D. from the Institute of Computing Technology, Chinese Academy of Sciences. Dr. Sun has led or executed 5 major funded research projects and published over 20 technical papers in journals and conferences related to computer network and distributed systems, including IEEE Transactions on Parallel and Distributed Systems (TPDS), IEEE Transactions on Mobile Computing (TMC), and IEEE International Conference on Computer Communications (INFOCOM). Her research interests include federated learning, edge intelligence, and privacy computing.
	
\end{IEEEbiography}

\begin{IEEEbiography}[{\includegraphics[width=1in,height=1.25in,clip,keepaspectratio]{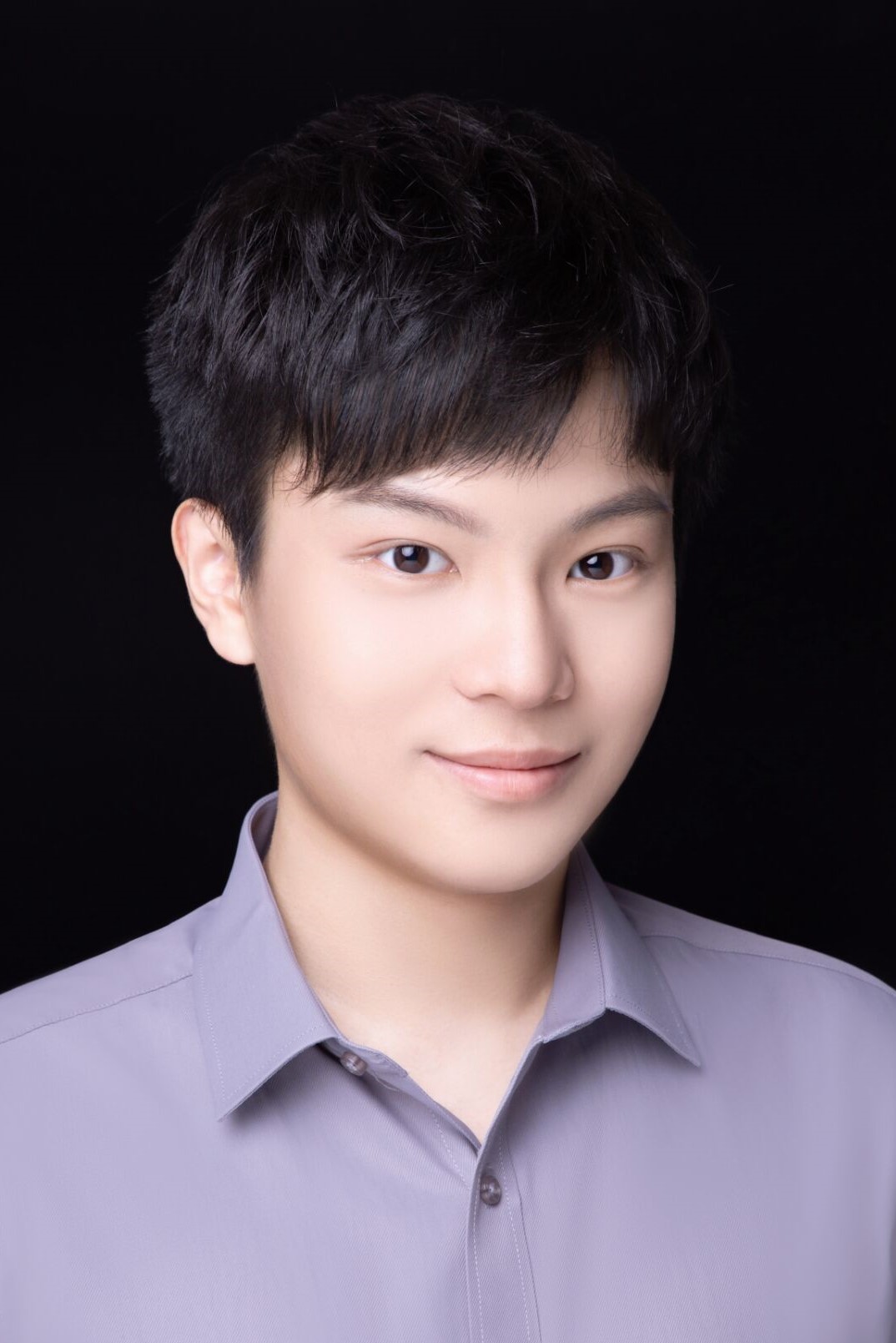}}]{Zhiyuan Wu}
	(Member, IEEE) is currently a research assistant with the Institute of Computing Technology, Chinese Academy of Sciences (ICT, CAS). He has contributed several technical papers to top-tier conferences and journals as the first author in the fields of computer architecture, computer networks, and intelligent systems, including IEEE Transactions on Parallel and Distributed Systems (TPDS), IEEE Transactions on Mobile Computing (TMC), IEEE International Conference on Computer Communications (INFOCOM), and ACM Transactions on Intelligent Systems and Technology (TIST). He has served as a technical program committee member or a reviewer for over 10 conferences and journals.  He is a member of IEEE, ACM, the China Computer Federation (CCF), and is granted the President Special Prize of ICT, CAS. His research interests include federated learning, mobile edge computing, and distributed systems.
\end{IEEEbiography}

\begin{IEEEbiography}[{\includegraphics[width=1in,height=1.25in,clip,keepaspectratio]{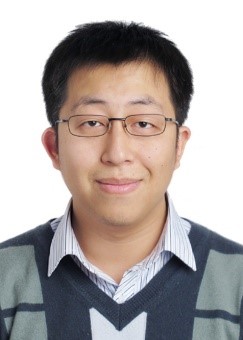}}]{Yuwei Wang}
(Member, IEEE) received his Ph.D. degree in computer science from the University of Chinese Academy of Sciences, Beijing, China. He is currently an associate professor at the Institute of Computing Technology, Chinese Academy of Sciences. He has been responsible for setting over 30 international and national standards, and also holds various positions in both international and national industrial standards development organizations (SDOs) as well as local research institutions, including the associate rapporteur at the ITU-T SG16 Q5, and the deputy director of China Communications Standards Association (CCSA) TC1 WG1. His current research interests include federated learning, mobile edge computing, and next-generation network architecture.
\end{IEEEbiography}

\begin{IEEEbiography}[{\includegraphics[width=1in,height=1.25in,clip,keepaspectratio]{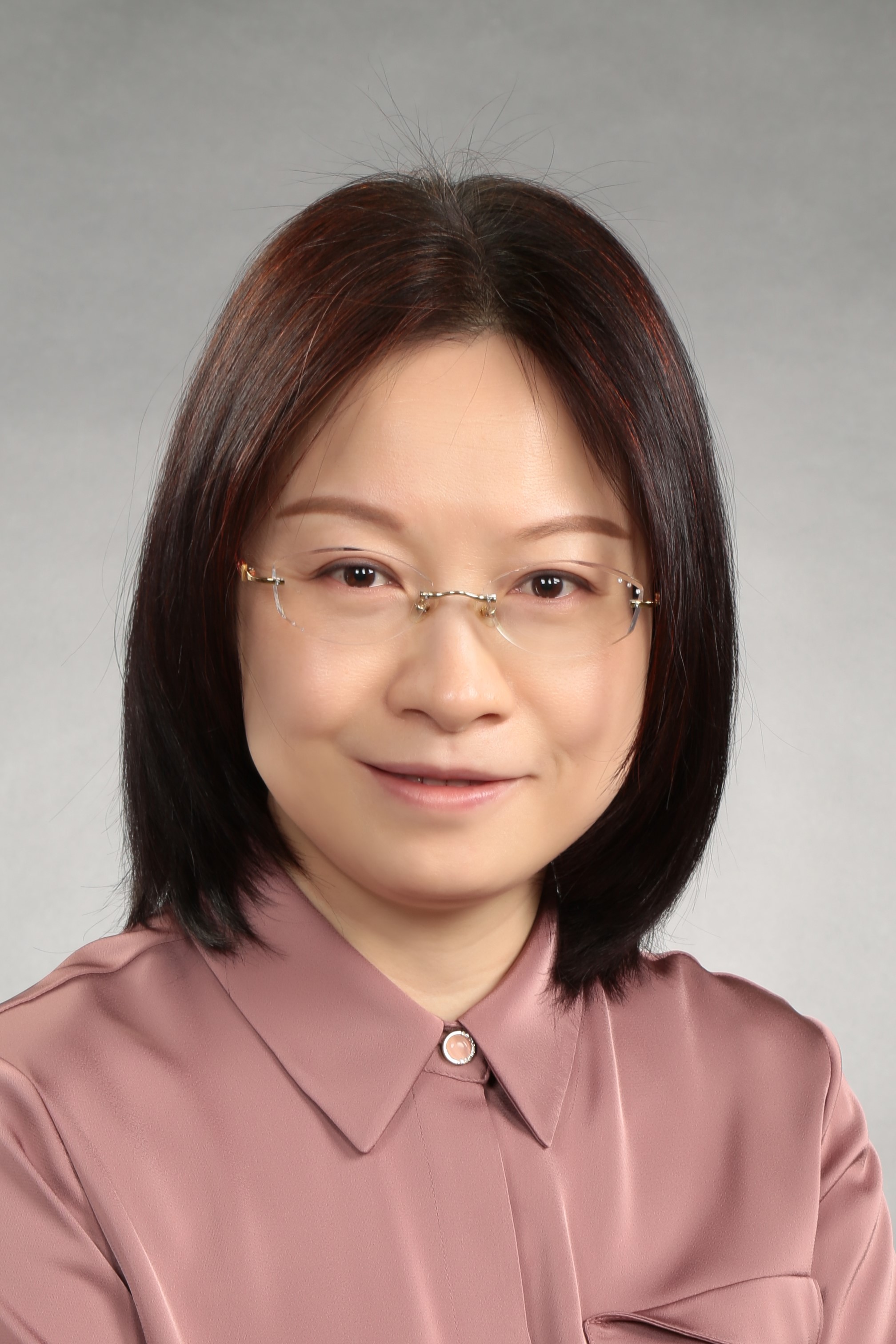}}]{Min Liu}
 	(Senior Member, IEEE) received her Ph.D degree in computer science from the Graduate University of the Chinese Academy of Sciences, China. Before that, she received her B.S. and M.S. degrees in computer science from Xi’an Jiaotong University, China. She is currently a professor at the Institute of Computing Technology, Chinese Academy of Sciences, and also holds a position at the Zhongguancun Laboratory. Her current research interests include mobile computing and edge intelligence.
 \end{IEEEbiography}

 \begin{IEEEbiography}[{\includegraphics[width=1in,height=1.25in,clip,keepaspectratio]{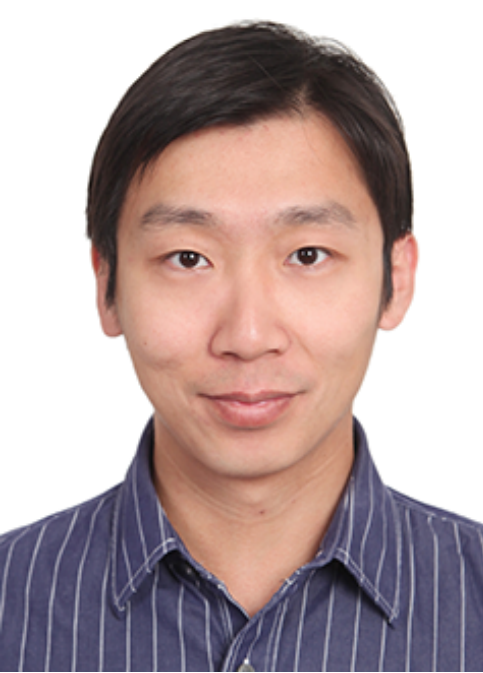}}]{Bo Gao} (Member, IEEE) 	received his M.S. degree in electrical engineering from the School of Electronic Information and Electrical Engineering at Shanghai Jiaotong University, Shanghai, China in 2009, and his Ph.D. degree in computer engineering from the Bradley Department of Electrical and Computer Engineering at Virginia Tech, Blacksburg, USA in 2014. He was an Assistant Professor with the Institute of Computing Technology at Chinese Academy of Sciences, Beijing, China from 2014 to 2017. He was a Visiting Researcher with the School of Computing and Communications at Lancaster University, Lancaster, UK from 2018 to 2019. He is currently an Associate Professor with the School of Computer and Information Technology at Beijing Jiaotong University, Beijing, China. He has directed a number of research projects sponsored by the National Natural Science Foundation of China (NSFC) or other funding agencies. He is a member of IEEE, ACM, and China Computer Federation (CCF). His research interests include wireless networking, mobile/edge computing, multiagent systems, and machine learning.
 \end{IEEEbiography}

\begin{IEEEbiography}[{\includegraphics[width=1in,height=1.25in,clip,keepaspectratio]{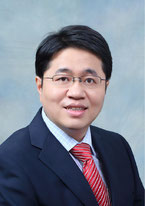}}]{Jingyuan Wang} received the PhD degree from the Department of Computer Science and Technology, Tsinghua University, Beijing, China. He is currently a professor of School of Computer Science and Engineering, Beihang University, Beijing, China. He is also the head of Beihang Interest Group on SmartCity, and vice director of the Beijing City Lab. He published more than 30 papers on top journals and conferences, as well as named inventor on several granted U.S. patents. His general area of research is data mining and machine learning, with special interests in smart cities.
 \end{IEEEbiography}

%\newpage
%\newpage
\clearpage
\appendices
\section*{Appendix}
%\renewcommand\thesection{\Alph{section}}
% \noindent
% {\Large \textbf{Appendix}}

    \begin{figure*}[b]
    	\centering
    	\includegraphics[width=0.9\textwidth]{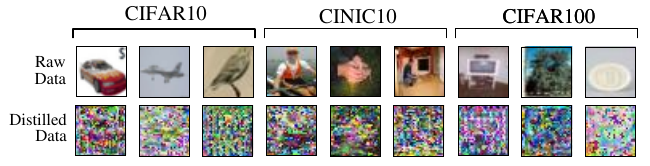}
    	\caption{Visualization of raw data and distilled data over image recognition datasets.}
            \label{data-distill}
    \end{figure*}
    
In the appendix, we provide additional details organized as follows:
\begin{itemize}
    \item 
    Section \ref{app-privacy} illustrates the privacy guarantees of the federated dataset distillation process. It compares raw and distilled data across three image recognition datasets, highlighting how the federated dataset distillation process obscures identifiable visual information to protect user privacy.
    \item 
    Section \ref{appendix-dataset} provides an overview of the tasks and datasets used in our experiments. It includes brief introductions of image recognition, audio understanding, mobile sensor data mining, and elaborate descriptions of CIFAR-10, CIFAR-100, CINIC-10, UrbanSound8K, and TMD datasets.
    \item 
    Section \ref{appendix-model-structure-configurations} describes the model structures used in our experiments, and provides detailed information of model parameters.
    %\item 
    %Section \ref{appendix-implementation-details} provides information on the model configurations for both homogeneous and heterogeneous settings, data partition strategies, and hyper-parameter settings across the experiments.
    \item 
    Section \ref{appendix-communication-computation} introduces the methodology for quantifying the communication costs of FL algorithms, including specific measures for different types of transmitted information like model weights, logits, sample index, and distilled data.
    %\item 
    % Section \ref{appendix-quantitive-communication} elaborates on detailed quantitative communication costs and efficiency speed-up ratios for image recognition tasks with different degrees of data heterogeneity.
    % \item 
    % Section \ref{appendix-results-downstream-tasks} provides additional evaluations of FedCache 2.0 on audio understanding and mobile sensor data mining tasks. It presents the average user model accuracy and communication efficiency for these tasks, demonstrating the effectiveness of FedCache 2.0 in diverse data modalities.
\end{itemize}

%\begin{refsection}

\section{Privacy Demonstration of Federated Dataset Distillation}
    \label{app-privacy}
    In this paper, federated data distillation is employed to develop abstract semantic representations on individual devices collaboratively. Figure \ref{data-distill} illustrates the comparison of raw local data and distilled data on three image recognition datasets. As shown in Figure \ref{data-distill}, raw datasets on each client feature rich and identifiable visual information, posing severe risks to user privacy. For instance, in the CIFAR-10 \cite{cifar10} dataset, the unprocessed images vividly display the characteristic features of certain common objects such as cars, birds, and plains. Similarly, in the CINIC-10 \cite{cinic10} dataset, the distinct contours of people are clearly discernible.
    On the contrary, the images become significantly obscured and unidentifiable after distillation. Specifically, the distilled data across all datasets exhibit a substantial reduction in distinct object contours and detailed features, preserving only rudimentary aspects of color and shape. While the transformed data preserves enough statistical attributes for effective local model training, it lacks the granularity necessary for image recovery. Hence, this transformation greatly enhances user privacy by significantly mitigating the risk of personal data leakage when such distilled data is processed centrally on servers.

\begin{figure*}[t]
        \centering
    \includegraphics[width=0.9\textwidth]{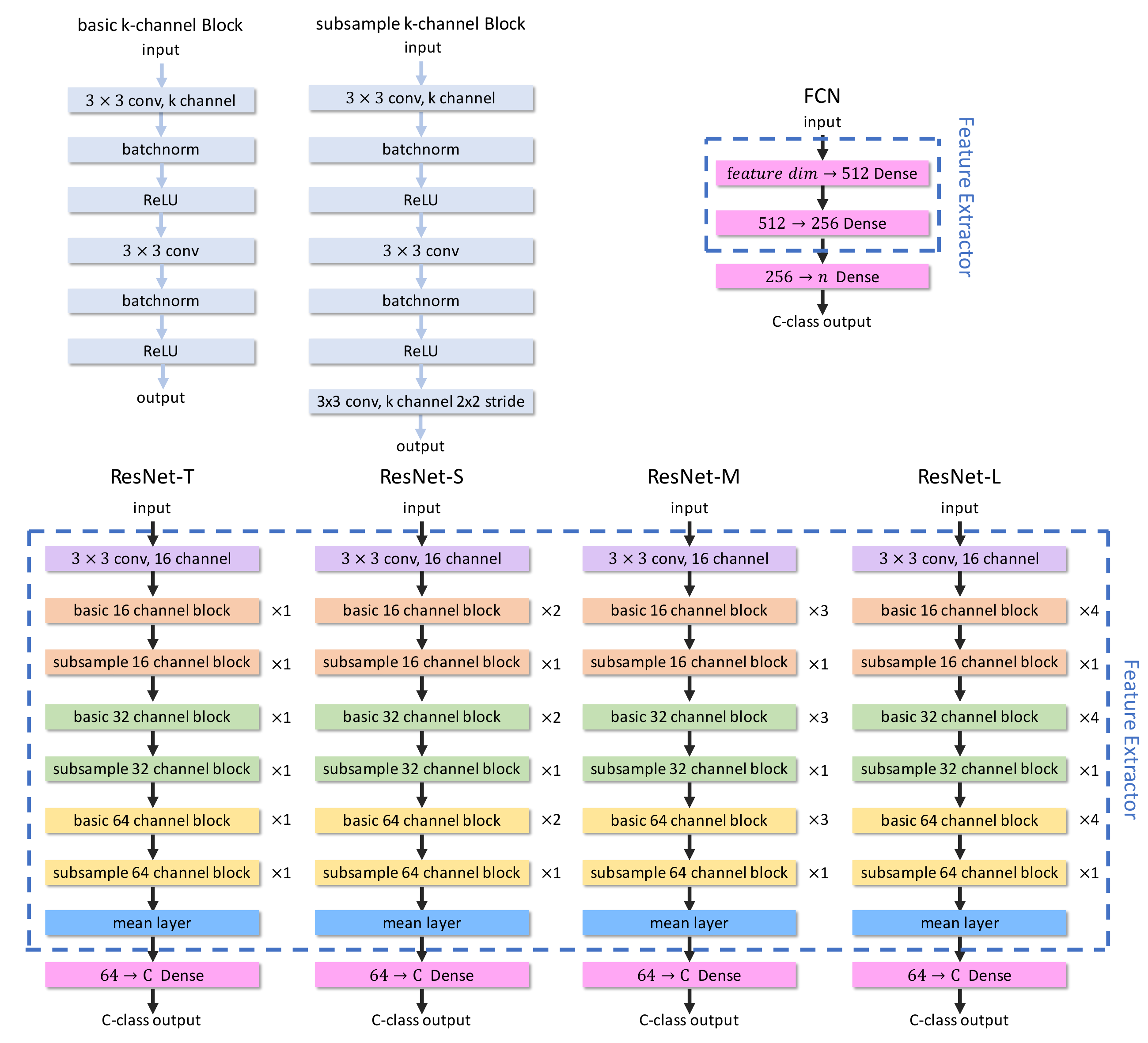}
        \caption{Detailed information of adopted model structures.}
        \label{model}
\end{figure*}

\section{Illustration of Tasks and Datasets}
\label{appendix-dataset}
In this section, we provide an overview of the tasks and datasets utilized in our experiments, which span three distinct domains: image recognition, audio understanding, and mobile sensor data mining.

\noindent
\textbf{Image Recognition.} Image recognition is a fundamental task in computer vision, involving the assignment of images into predefined classes based on their visual content, such as color, texture, and shape. In our experiments, we evaluate our method on three image recognition datasets, which are CIFAR-10, CIFAR-100 \cite{cifar10}, and CINIC-10 \cite{cinic10}. All of these datasets consist of common objects in daily life, represented by vehicles, animals, and humans. They offer a diverse and challenging set of images that test the robustness and precision of FL algorithms in handling visual data.

\begin{table*}[t]
        \centering
        \renewcommand{\arraystretch}{1.2}
        \caption{Number of parameters of adopted models.}
        \label{model-params} 
        \begin{tabular}{l|cccccc}
            \hline
            \textbf{Model}      & \textbf{ResNet-T} & \textbf{ResNet-S} & \textbf{ResNet-M} & \textbf{ResNet-L} & \textbf{FCN-U} & \textbf{FCN-T} \\
            \hline
            \textbf{Num. Params.} & 171.0K           & 265.9K           & 360.8K           & 455.8K           & 151.3K & 162.5K \\
            \hline
        \end{tabular}
\end{table*}

\noindent
\textbf{Audio Understanding.} Audio understanding involves classifying audio clips into various categories based on their acoustic properties. We employ the UrbanSound8K dataset \cite{urbansound} in our experiments, which contains labeled sound excerpts from 10 distinct classes, including car horns, street music, and children playing. Each sound clip is less than four seconds long, providing a compact yet comprehensive acoustic profile for analysis. This dataset is particularly valuable for evaluating the performance of audio recognition FL systems in naturalistic urban environments.

\noindent
\textbf{Mobile Sensor Data Mining.} Mobile sensor data mining focuses on analyzing data from mobile device sensors to infer user activities or contextual settings. In our experiments, we utilize the TMD dataset \cite{carpineti2018custom} designed to identify transportation modes based on smartphone sensor readings. The TMD dataset is ideal for assessing the capability of FL algorithms to process and interpret complex, real-time sensor data in the context of mobile computing.

\section{Detailed Information on Model Structures}
    \label{appendix-model-structure-configurations}
    In this section, we provide detailed descriptions of the model structures used in our experiments. For experiments on image recognition, we employ deep residual network (ResNet) \cite{he2016deep} with different numbers of layers to represent models with different structures, which are ResNet-T, ResNet-S, ResNet-M, and ResNet-T. For experiments on audio understanding as well as mobile sensor data mining, we adopt fully connected networks (FCN) that satisfy task-specific inputs and outputs sizes. We provide detailed information on model structures in Figure \ref{model}, and the number of parameters of adopted models in Table \ref{model-params}.
    
    %Table \ref{model}

% \begin{figure}[t!]
% 	\centering	\includegraphics[width=1.0\textwidth]{model.pdf}
% 	\caption{Detailed information of adopted model structures.}
%         \label{model}
% %\end{figure}

% \vspace{15pt}

% %\begin{figure}[h]
%     	\centering
%     	\includegraphics[width=1.0\textwidth]{data-dist.png}
%     	\caption{???}
%             \label{appendix-data-dist}
% \end{figure}

% \begin{table}[t]

% \caption{Number of parameters of adopted models.}
% \label{model-params}
% \centering
% \begin{tabular}{l|cccccc}
% \hline
% \textbf{Model}      & \textbf{ResNet-T} & \textbf{ResNet-S} & \textbf{ResNet-M} & \textbf{ResNet-L} & \textbf{FCN-U} & \textbf{FCN-T}

% \\ \hline
% \textbf{Num. Params.} & 171.0K           & 265.9K           & 360.8K           & 455.8K           & 39.9K & 41.8K \\ \hline
% \end{tabular}
% \end{table}

    \section{Communication Cost Calculation}
    \label{appendix-communication-computation}
    In this section, introduce how communication cost is computed in our paper. We quantify the communication cost of all considered FL algorithms in terms of pure information transmission- specifically model weights, logits, sample index, and distilled data—between clients and the server. Our communication calculation methodology leverages standard units of bytes to measure communication costs, ensuring precise and scalable metrics. For a comprehensive evaluation, we compute the communication cost of all considered algorithms where the overheads vary based on the type and frequency of transmitted information:

    \noindent
    \textbf{MTFL \cite{mills2021multi}.} Both model and optimizer parameters are serialized and transmitted in each communication round. All parameters are encoded as tensors in the float format, with each element occupying 4 bytes.

    \noindent
    \textbf{kNN-Per \cite{marfoq2022personalized} and SCDPFL \cite{chen2024spectral}.} Similar to MTFL, these methods involve transmitting model parameters in float tensors in each communication round.
    
    \noindent
    \textbf{FedKD \cite{wu2022communication}.} Parameters of the student model (ResNet-T) are transmitted in each round.

    \noindent
    \textbf{FedCache \cite{wu2024fedcache}.} This method introduces additional complexity by transmitting sample hashes (in float format), sample index (in integer format), and logits (in float format). 

    \noindent
    \textbf{FedCache 2.0.} This method extends FedCache by incorporating distilled data, which is uploaded and downloaded each round. Each client uploads one synthetic data sample per class, with the volume of downloaded data regulated by the hyper-parameter $\tau$. Distilled images are initially processed into a $32\times32\times3$ array in uint8 format (1 byte each) before conversion to JPG, optimizing communication cost.

\ifCLASSOPTIONcaptionsoff
  \newpage
\fi

% \vfill
% \newpage
% \appendices
% \onecolumn
 \end{document}